\documentclass[journal]{vgtc}                     

\vgtccategory{Research}

\title{Beyond One Output: Visualizing and Comparing Distributions of Language Model Generations}

\author{%
  \authororcid{Emily Reif}{0000-0003-3572-6234},
  \authororcid{Claire Yang}{0009-0003-6170-4183},
  \authororcid{Jared Hwang}{0009-0000-4769-1521},
  Deniz Nazar,
  \authororcid{Noah A. Smith}{0000-0002-2310-6380}, and
  \authororcid{Jeff Heer}{0000-0002-6175-1655}
}

\authorfooter{
  \item Emily Reif, Claire Yang, Jared Hwang, Deniz Nazar, Noah Smith, and Jeff Heer are with the University of Washington, Seattle, WA, USA. Noah Smith is also with the Allen Institute for AI.
  \item E-mail: \{emreif,claireyy,jaredhwa,denizn,nasmith,jheer\}@cs.washington.edu
}

\abstract{%
  Users typically interact with and evaluate language models (LMs) via single outputs, where each output is just one sample from a broad distribution of possible completions.
  This interaction hides distributional structure such as modes, uncommon edge cases, and sensitivity to small prompt changes, leading users to over-generalize from anecdotes when iterating on prompts for open-ended tasks.
  We conduct a formative study with researchers who use LMs ($n{=}13$) examining when stochasticity matters in practice, how they reason about distributions over language, and where current workflows break down. Informed by this study, we introduce \tool{}, an interactive visualization that represents multiple LM generations as overlapping paths through a text graph, revealing shared structure, branching points, and clusters while preserving access to raw outputs.
  We evaluate across three crowdsourced user studies ($N{=}47$, $44$, and $40$ participants) targeting complementary distributional tasks.
  Our results support a hybrid workflow: graph summaries improve structural judgments (e.g., assessing LM output diversity), while direct output inspection remains more effective for detail-oriented tasks.
}


\teaser{
  \centering
  \includegraphics[width=\linewidth, alt={Graph visualization of LM output distributions.}]{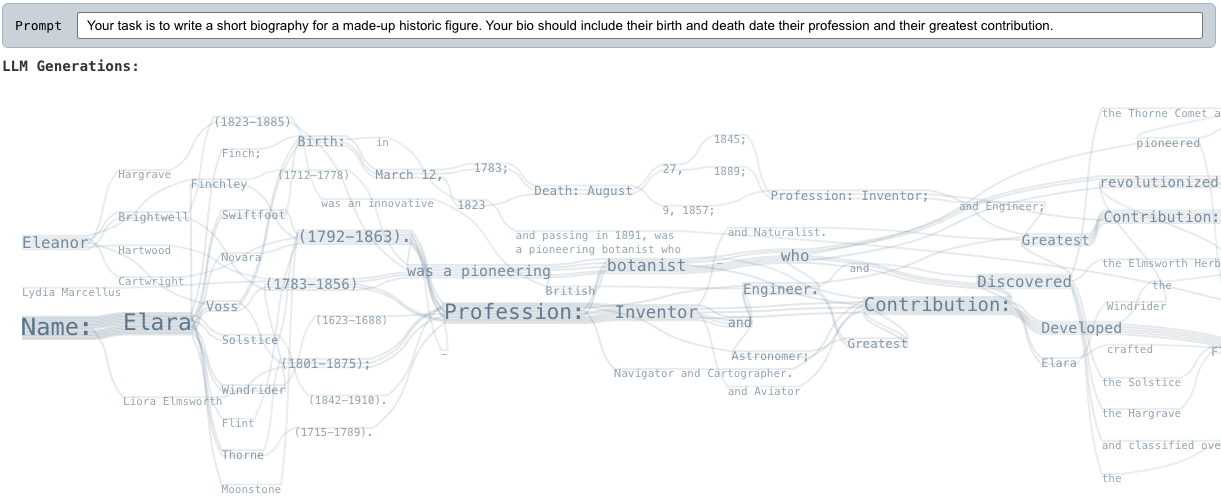}
  \caption{%
    \tool{} visualizes outputs from one or more prompts or model configurations. Here we see a distribution of generations for the same ``made up historic figure'' request. While the request is open-ended, many results are unexpectedly similar, both semantically (most of the characters are named ``Elara'') and structurally (most of the generations share the same syntactic template).%
  }
  \label{fig:teaser}
}

\renewcommand{\manuscriptnotetxt}{}

\graphicspath{{figs/}{figures/}{pictures/}{images/}{./}}

\usepackage{algorithm}
\usepackage{algpseudocode}
\usepackage{tabu}
\usepackage{array}
\usepackage{booktabs}
\usepackage{colortbl}
\usepackage{ccicons}
\usepackage{mathptmx}
\usepackage{subcaption}
\usepackage{placeins}
\usepackage{float}

\newcommand\tool[1]{GROVE}

\newcommand\edit[1]{{#1}}

\begin{document}

\maketitle

\section{Introduction}

Language models (LMs) exhibit a range of distributional quirks in how they generate outputs. For example, when prompted to invent a fictional character, the name ``Elara Voss'' appears far more frequently than one might expect \cite{MaxRead, wattenberg_elara_2025}, to the point where it is now cited as a sign of AI-generated text \cite{wiki:Signs_of_AI_writing}. This is particularly striking given that the name has no real-world provenance: prior to 2023, ``Elara Voss'' did not appear in public records or published text \cite{MaxRead}. Despite this, it now occupies a high-probability region of many models' distributions. \edit{Fig.~\ref{fig:teaser}} illustrates this effect.
This is not an isolated case, as LMs generate homogeneous responses to open-ended prompts \cite{jiang2025artificialhivemindopenendedhomogeneity}, are prone to mode collapse \cite{zhang2025noveltybench}, and struggle with random number generation \cite{hopkins2023can}. LM inconsistency may manifest in non-humanlike \cite{reinhart2025LMs} and difficult-to-predict ways, and can contribute to cultural homogenization \cite{10.1145/3706598.3713564}. As users typically see only one response, these issues can be hard to identify.

There are consequences to this stochasticity (or lack thereof). Swoopes et al.~\cite{Swoopes2025stochasticity} show that presenting a single response fosters undue trust and anthropomorphization, whereas showing multiple responses helps users understand the model's probabilistic nature. Similarly, when iterating on prompts, Zamfirescu-Pereira et al.~\cite{10.1145/3544548.3581388} found that users routinely ``over-generalized from single data points,'' either stopping when a desired behavior appeared once, or abandoning a promising prompt after one failure. Sampling the same prompt multiple times may produce distinct outputs, making it difficult to know whether observed differences reflect the prompts or simply randomness. 

These problems are especially acute in LM-supported open-ended tasks \cite{10.1145/3626772.3657914}, which include scientific discovery \cite{zhang2025exploring}, mental health support \cite{Ma2023UnderstandingTB}, and agentic AI \cite{acharya2025agentic}. In these tasks, the user must judge the LM's accuracy on open-ended behavior from sparse feedback. To better surface LLM variability and support distributional sensemaking, Gero et al.~\cite{10.1145/3613904.3642139} argue for ``mesoscale'' inspection (tens to hundreds of outputs to a single prompt). We build on that motivation with a complementary visualization approach. We conduct a formative study (\S\ref{sec:formative}) with researchers using LMs ($N{=}13$), focusing on how they reason about distributions over text. We investigate when stochasticity matters, what concepts they use (central tendency, spread, modes, outliers), and where workflows break down. Participants described LMs as infrastructural material, ``filling the gaps'' (P5) where custom models or crowdwork would be impractical, yet they lack tools to reason about the underlying distribution. Single-output interfaces hide this distribution, but building a tool is challenging because text has no agreed-upon units of variation.

Based on findings from our formative study, we designed a \emph{G}raph \emph{R}epresentation of \emph{O}utput \emph{V}ariability and \emph{E}xamples (\tool{}), and introduce it in \S\ref{sec:design}\footnote{See demo \href{https://emilyreif.com/llm-consistency-vis/}{website.}}. \tool{} is a graph-based visualization that merges overlapping \edit{model outputs} into a single structure. Each output is represented as a path through this graph, enabling a spatial visualization of branching, convergence, and frequency of sampled outputs. We conduct three crowdsourced evaluation experiments comparing this graph-based visualization to a list view, finding that different tasks necessitate different interfaces. In comparing the diversity of two output distributions, participants had higher accuracy and strongly preferred the graph-based visualization. However, when identifying more specific differences between two output distributions or comprehending a single output distribution, participants exhibited higher accuracy using the list view.
Overall, participants stated that they favored a combined interface, where the graph is useful for summarizing the high-level distribution, and a list of raw outputs is helpful for inspecting details. \edit{Our evaluation suggests that \tool{}'s graph encoding is most effective for relatively short, structurally aligned generations where divergence is localized (settings where shared paths branch and reconverge in interpretable ways).}
Our contributions are as follows:
\begin{itemize}
    \item \textbf{A characterization} of when stochasticity matters, how experts reason about ``distributions'' over language, and where workflows break down, along with \textbf{design needs} for sensemaking of the distributional structure of natural language.
    \item \textbf{\tool{}, a graph-based visualization} that adapts prior text graph techniques to LM generations, revealing shared structure, branching, clusters, and outliers. We integrate this into a tool supporting comparison across prompts, models, and decodings.
    \item \textbf{Empirical insights from three evaluative user studies} characterizing how interface design affects users' ability to reason about LM output distributions. The graph outperforms lists on diversity comparison (accuracy, time, preference); lists outperform the graph on fine-grained two-way- and single-distribution tasks.
\end{itemize}

\section{Related Work}

\subsection{Quantitative measures of LLM diversity from NLP}
NLP research provides quantitative measures of LM variability.
Kuhn et al.~\cite{kuhn2023semantic} quantify semantic uncertainty across sets of LM outputs, while
NoveltyBench~\cite{Zhang2025-sw} and Jiang et al.~\cite{jiang2025artificialhivemindopenendedhomogeneity} both present datasets of open-ended prompts and their outputs, finding substantial repetition both within a single model and across models.
Related work attributes diversity loss during instruction tuning~\cite{o'mahony2024attributing}, proposes methods that encourage more diverse output distributions~\cite{Zhang2024-lx}, and detects narrative collapse~\cite{Hamilton2024-sl}.
These aggregate metrics are useful for capturing overall trends and guiding large-scale training, but they rarely ground a practitioner’s judgment on \emph{their} specific samples. When used during prompt iteration, a single score obscures the underlying structure of the outputs (i.e., how the lack of variability actually manifests). Our work complements these metrics by making branching, convergence, and repetition directly visible in a user’s own generations.

\subsection{Sensemaking of LM outputs}
An exciting new line of HCI and visualization research offers many ways to support sensemaking of LM outputs. Some visualizations focus on a single (possibly long) output, like ReasonGraph~\cite{Li2025-dl} and Interactive Reasoning~\cite{pang2025interactive}. Others, like LMdiff~\cite{Strobelt2021-ag} and LLM Comparator~\cite{Kahng2024-mm} help analysts contrast models at a global level by inspecting token- or example-level differences across datasets.

\textbf{Difficulty of prompt iteration.}
Prior work also shows that experts and non-experts struggle to iterate on prompts because outputs are brittle and hard to predict~\cite{10.1145/3563657.3596138,10.1145/3544548.3581388}; designers in one study likened working with LMs to ``herding cats''~\cite{10.1145/3563657.3596138}.
Prompting in the Dark~\cite{He2025-ij} further finds that users often fail to improve LM performance over multiple rounds without ground-truth labels.
Together, these results motivate interfaces that make \emph{how prompts change output distributions} visible, not only a single sample.
Systems such as ChainForge~\cite{Arawjo2024-rw} and AI chains~\cite{wu2022ai} support structured prompt testing and chaining; we complement that thread by foregrounding distributional change when the same prompt is sampled many times.

\textbf{Many outputs from one prompt.}
Gero et al.~\cite{10.1145/3613904.3642139} make the case for ``mesoscale'' inspection, showing tens to hundreds of completions, and show that surfacing cross-output structure (exact matches, positional similarity) helps analysts reason about a sample.
We share that aim but pursue a complementary encoding: rather than showing every response in full text, we merge tokens into one graph so branching, reconvergence, and frequency are visible. 
Swoopes et al.~\cite{Swoopes2025stochasticity} examine how presenting many answers at once affects trust, anthropomorphization, and workload, and evaluate color-coded span highlighting as cognitive support; their results show that layout and highlighting choices shape how users interpret diversity, above and beyond whether more than one response is shown.
Brath et al.~\cite{brathvisualizing} also summarize many outputs to the same prompt, this time using mind maps, targeted highlighting, and clustering;  a related study~\cite{brath2023visualizing} traces how distributions drift under different style transforms. Sevastjanova et al.~\cite{Sevastjanova2023-lq} also visualize many outputs using clusters and embedding visualizations. Spinner et al.~\cite{10.1145/3652028} visualize the beam search algorithm of generation directly to show the distribution, though their encoding (a tree) and goal (interacting with and editing the generation) differ.
RELIC~\cite{Cheng2024-xs} looks at factual consistency across multiple sampled generations.
Together these systems share our interest in multiplicity. We contribute a merged token graph oriented to open-ended text and report which distributional judgments it supports relative to list-based baselines.

\subsection{Graph views and sensemaking over text corpora.}
Our graph encoding sits in a longer tradition of sensemaking for unstructured text~\cite{pirolli2005sensemaking} and particularly of Shneiderman's ``overview first, zoom and filter, then details-on-demand'' guidance~\cite{shneiderman1996eyes}.
Much of this prior work does not target LM outputs, but other types of document collections: WordTree~\cite{wattenberg2008word} and WordGraph~\cite{riehmann2012wordgraph} contextualize search results by showing neighboring words and local structure. PhraseNet~\cite{van-Ham2009-me} graphs words linked by a template relation, and AbstractExplorer~\cite{Gu2025-bf} combines role highlighting with alignment to support comparative reading across many abstracts.
Token graph visualizations include SentenTree~\cite{hu2016visualizing}, which aggregates frequent sequential patterns in tweets, and J\"{a}nicke et al.~\cite{janicke2014visualizations}, who compare re-used passages of text with graph-based alignments.

A related line of work in machine translation provides tools for visualizing uncertainty in word lattices~\cite{collins2007visualization,collins2006leveraging}. Systems like NMTVis~\cite{Munz2022-bg} further coordinate tree, graph, and metric views to support inspection of variation across translations. However, these approaches assume a tightly structured space anchored to one source sentence to be translated. In contrast, we focus on open-ended LM sampling, where outputs are less constrained and predictable.

In conclusion, related work motivates both measuring and seeing distributional behavior.
We contribute a graph-based visualization tuned to LM output sets, empirical evidence on which distributional tasks it supports relative to raw lists, and a design that prioritizes within-prompt structure alongside comparison across prompts.

\section{Formative study}
\label{sec:formative}
To better understand how LM researchers and practitioners currently assess, address, and leverage the stochasticity of models, we conducted a semi-structured interview study. We wanted to answer four main questions:
\begin{enumerate}
    \item When does it matter that LMs are stochastic, i.e., that they produce a distribution of responses rather than just one?
    \item Given this stochasticity, how do people iterate on LM-based solutions for open-ended tasks with no ground truth?
    \item  When is it desirable to have a wide distribution of outputs?
    \item How do they operationalize the notion of a ``distribution'' in their contexts, given the unstructured nature of natural language?
\end{enumerate}

\subsection{Method}
We recruited thirteen researchers via snowball sampling, including researchers in NLP and HCI with experience using LMs for open-ended tasks; we excluded those without hands-on experience evaluating LM outputs. These included ten PhD students, two industry research scientists, and one faculty. This study was approved by the IRB of the University of Washington (No. STUDY00024291). Participants provided informed consent before participating.

\subsection{Procedure}
Each session followed a semi-structured protocol lasting 30-45 minutes (for the full protocol, see \S\ref{sec:protocol}):

\begin{enumerate}
    \item \textbf{Introduction (2 to 3 min)} Study purpose and scope.
    \item \textbf{Background (20 min)} Current research focus, LM workflows, when multiple outputs matter, and how they examine many outputs at once.
    \item \textbf{Prototype demo (10 min)} Walkthrough of an early prototype that included a version of the graph, designed to elicit feedback on what aspects of this encoding could be useful or confusing to participants. The demo also included 5 minutes of feedback.
    \item \textbf{Use case brainstorming (5 min)} Hypothetical integration into their work and potential extensions.
    \item \textbf{Wrap-up (2 min)} Final thoughts.
\end{enumerate}

\subsection{Participant Use of LMs}
Participants used LMs across a range of open-ended and applied tasks, including generating synthetic datasets (e.g., movie reviews or reasoning traces), prototyping and evaluating systems (e.g., coding assistance or alignment pipelines), and supporting domain-specific applications such as empathetic communication, medical question answering, and story generation. These workflows often involved repeatedly sampling multiple outputs to assess properties like diversity, correctness, or style.

\subsection{Analysis}
We analyzed the interviews using an iterative, inductive approach. One author conducted open coding over multiple passes to identify recurring themes, revising and consolidating codes as patterns emerged. Preliminary themes were discussed and refined in collaboration with coauthors to ensure they reflected patterns observed across participants. Our analysis was informed by established approaches to thematic analysis~\cite{morseCriticalAnalysisStrategies2015, braunUsingThematicAnalysis2006}, but focused on surfacing common patterns in how participants reason about and work with LM output variability rather than constructing a comprehensive codebook or exhaustive taxonomy.

\subsection{Results/Findings}
Across interviews, participants emphasized that evaluating LM behavior on novel, open-ended tasks is fundamentally a distributional problem: single outputs are misleading, automated metrics often fall short, and practitioners are left stitching together qualitative judgments across many examples.
\edit{Below, we trace how practitioners use LMs in practice, what they need to reason about output distributions, and how those needs map to design goals for distributional visualization.}

\edit{\subsubsection{Adoption without evaluation: LMs as infrastructure}}

Reinforcing findings from related work \cite{wu2022ai}, we find that participants increasingly treat LMs as infrastructural material, i.e., flexible components embedded within larger systems to handle subtasks like summarization, annotation, or content generation. 
This isn't necessarily out of enthusiasm for LMs; in many cases, LMs are often the only practical way to achieve a fluent natural language interface with reasonable accuracy on novel tasks, without large-scale data collection, custom training, or reliance on crowd workers. As P6 said,``there just wasn't really anything else that would have worked [for our task]''. However, even though the LM can hypothetically fill the gap, participants lamented that there is no straightforward way to evaluate how well these models are actually performing. 

\edit{\subsubsection{LMs enable nuanced tasks, and nuanced failures}}
Prior evaluation methods like accuracy on pre-existing benchmarks, or automatic diversity scores are not enough to determine whether an LM will work in a user's specific context. 
When evaluating prompts, 61\% of the participants said they often start with automatic metrics, but (P4): ``quantitative evaluation isn't great, [they don't] apply to our problem.''
Similarly, P5 asks, ``How do we quantify it? For example, we can use coherence, but that's limited. Empathetic reaction? How do we measure these nuanced things?''

On the other hand, in some cases automatic metrics \textbf{are} enough, especially when working on general model optimization rather than using it for a specific task. When working on a project around hallucinations, P9 did a few tests of the automatic metric, and ``it seemed OK so I didn't look too closely.'' P2 also used standard benchmarks when improving general model accuracy.

In general, though, sensemaking (looking directly at the outputs to get a sense of whether they ``seem good enough'') is the practical default. Typically, this inspection took place in IPython notebooks or raw text files. Participants emphasized that this lightweight and qualitative mode of iteration is often more productive than trying to formalize a benchmark too early in the design process.

\edit{\subsubsection{Practitioners reason over output distributions, not single completions}}
Participants noted that examining a single output can create a misleading impression of model behavior. They therefore evaluate prompts by looking across many outputs, despite the effort required. Even before API access, P1 recruited her husband and a graduate student to manually run prompts hundreds of times in the ChatGPT UI to surface long-tail behaviors. P13 echoed this difficulty, observing that ``it's a huge problem [when iterating on prompts] that a single output from model A could have also come from model B.'' There were two main types of diversity that were important to the participants during prompt iteration:
\begin{itemize}
    \item \textbf{Within-input diversity}: Given the same input, how varied are the possible outputs? Some tasks required consistency (e.g., facts), while others benefited from diversity (e.g., creative writing).
    \item \textbf{Diversity across prompts}: When iterating on a prompt, are the differences in outputs that we observe meaningful or just noise? 
\end{itemize}

Overall, participants stated that they want to reason about distributions of behavior, not isolated outputs. 

\edit{\subsubsection{What ``distribution'' means for natural language}}
Although participants routinely talked about ``the distribution'', ``diversity'', or ``similarity'' of model outputs, there was no shared operationalization for text. This ambiguity reflects the lack of reliable automated metrics. Unlike numeric or categorical outputs, natural language has no agreed-upon units of variation, and no single measure that captures the multiple dimensions along which text can differ. Instead, this was an ``impression test'' (P3) that users got from directly looking at outputs. How many outputs participants needed to judge these properties varied from a handful to hundreds. 

Outputs differ along multiple axes: participants pointed to variation in style or tone, character names, or narrative strategy. For example, P1 noticed that when prompting the model to write a story about a boy and his pet, ``there might be many different plotlines but they all are about a boy named Joe and a dog named Fido. Is that similar?''

But even if it is hard to operationalize, participants did care about several aspects of the distribution that have statistical analogs. These included \textbf{the average output} or central tendency of the model, \textbf{the spread or deviation}, once they determined which features actually mattered in their task, \textbf{where the outputs diverged}, and \textbf{outliers}, which were rare but disproportionately important for reliability and safety.

Overall, participants treated distributional reasoning for text as inherently qualitative. In other words, understanding what varies and what should vary requires human interpretation before any automatic evaluation can be meaningfully applied.

\edit{\subsubsection{Inspecting every output directly does not scale}}
Participants routinely sample multiple outputs (often five to ten) to ``get a feel'' for this distribution. But it was overwhelming to look at a giant block of text that results when listing the outputs, and they found it hard to reason about structure or overlap (i.e., what was shared, unique, or missing between outputs).
P11 stated that ``It would be useful for me to incorporate multiple examples into my workflow, but it's expensive to generate, and hard to qualitatively understand''. In general, looking at the data was often overwhelming and felt too subjective.

\edit{\subsubsection{Variation beyond evaluation}}
\edit{Beyond evaluation,} there were a broad range of use cases where the range and diversity of potential outputs mattered.

 \textbf{Inherently open-ended prompts.} P1 and P10 were both using LMs in the creative writing context, where it was desirable to give the user a range of potential story directions.

\textbf{Synthetic data.} Many participants (P3, P5, P6, P8, and P11) used LMs to generate data. The process was generally similar: give the LM an open-ended prompt, e.g., ``generate me a movie review'', and call it many times to create the dataset. It was a huge issue if the generated data was too uniform.

\textbf{Matching human distributions.} P13 cared ``that the distribution isn't static, and matches the output of actual people.'' Relatedly, P12 noted that seeing the distribution is `` useful for AI literacy-- if I do ``picture of man'' is it always a white guy with short hair?''

\textbf{Reasoning traces.} Reasoning traces are often designed to be divergent, especially in ensembled systems. 

\textbf{Explicitly creating diversity} Several participants intentionally built systems that explicitly elicit diversity: P4 who was building a research tool with a set of simulated (LM) experts from different scientific domains, asked ``If I've accidentally created the same agent multiple times, or they're all giving the same advice, then why not just make one?''. Similarly, P9 created a diversity of chess agents with different strategies. Both of them wanted to see how the different agents diverged, and in what ways. 

\edit{Reliability concerns were equally salient, especially for participants building complex end-user-facing systems, where randomness was a liability.} 
Importantly, participants did not equate consistency with identical outputs. Instead, they wanted to be able to control the variation, maintaining semantic or logical consistency while allowing flexibility in style or phrasing (P3). This highlights the challenge of defining and measuring consistency in natural language, where acceptable variation depends on what aspects of the output are contextually important.

\edit{\subsubsection{Design implications}}
\edit{Together, these findings describe a recurring workflow, where practitioners adopt LMs despite weak evaluation (\S3.5.1), but standard metrics fail for open-ended tasks. Instead, they inspect many outputs directly (\S3.5.2), reasoning about the distribution of these outputs, not single completions (\S3.5.3), in terms of overlap, modes, spread, and outliers (\S3.5.4). However, reading every output does not scale (\S3.5.5), and the same reasoning applies beyond evaluation across tasks that differ in what should stay stable or vary (\S3.5.6). Current tools mostly expose one completion or unstructured lists, leaving a gap between how experts think about LM behavior and what interfaces support. We need tools that summarize variation across many outputs without sacrificing raw-text access. We next formalize these needs as design goals.}

\section{Design Goals}

\begin{figure*}
  \centering
  \includegraphics[width=\textwidth]{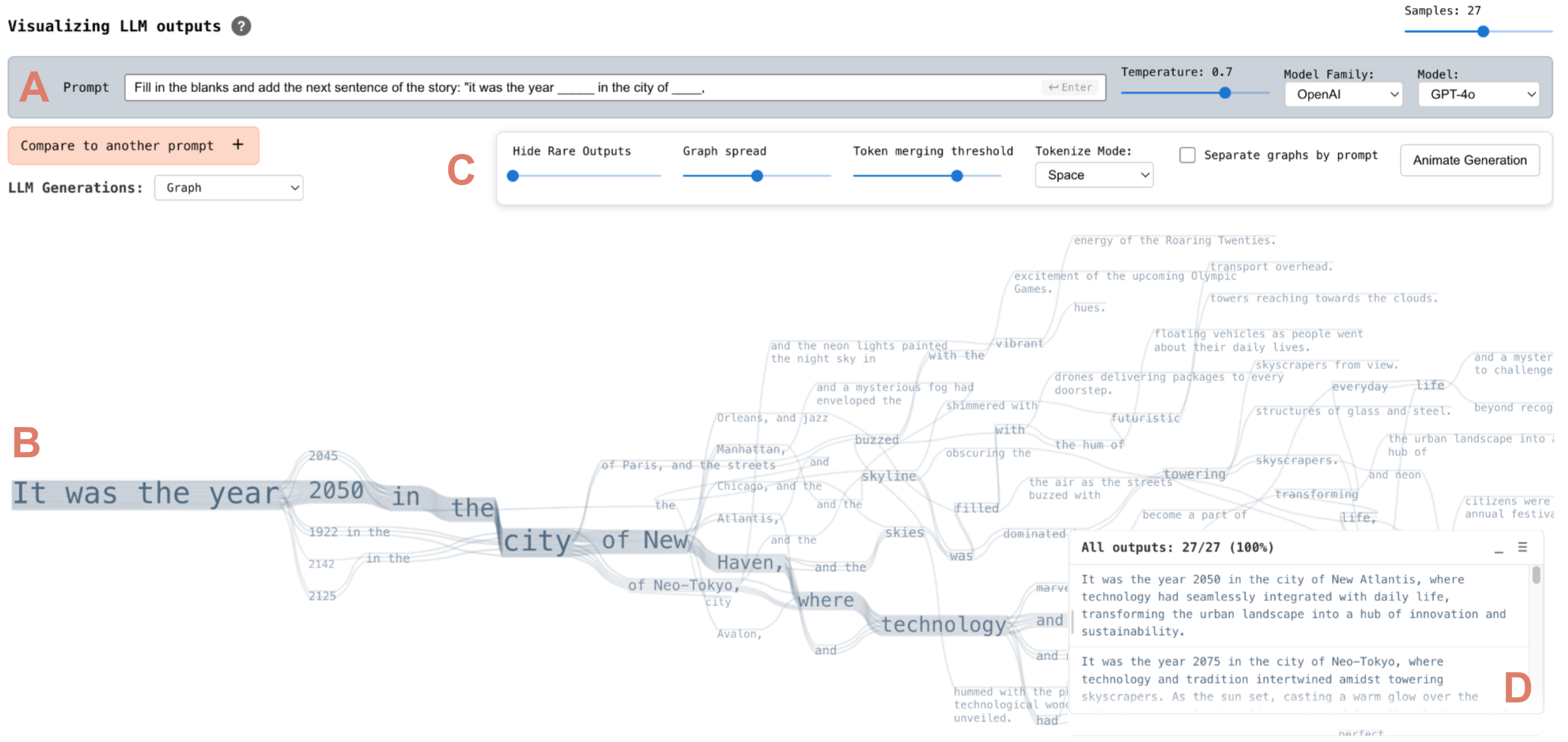}
  \caption{The basic \tool{} interface. \textbf{A:} the global controls for generation, including the prompt, number of generations, model, and temperature. \textbf{B:} the graph of generations. \textbf{C:} the graph controls, including various ways to simplify and render it. \textbf{D:} The original raw text outputs, in list form, which can be expanded or minimized.}
    \label{fig:basic}
\end{figure*}

\edit{Guided by the formative workflow above, we derive six design goals (DG) for a distributional visualization tool:}
\begin{description}
    \item[(DG1)]\label{dg:1}\edit{Provide a compact visual summary of many outputs without requiring users to read each one, conveying the overall ``shape'' of the response space.}
    \item[(DG2)]\label{dg:2}\edit{Surface what outputs share: repeated words, phrases, entities, and themes.}
    \item[(DG3)]\label{dg:3}\edit{Show where those outputs diverge: branching and reconvergence, outliers, and artifacts.}
    \item[(DG4)]\label{dg:4}\edit{Support comparison across inputs: how shared structure, divergence, and overall shape differ between output sets from different prompts, models, or decoding settings.}
    \item[(DG5)]\label{dg:5}Preserve access to raw text.
    \item[(DG6)]\label{dg:6}Enable rapid, interactive hypothesis testing by letting users modify prompts and immediately observe changes in the output distribution.
\end{description}
\edit{\noindent These goals map to the formative findings. DG1--DG3 address distributional reasoning for a given input (\S3.5.3--3.5.5): summarizing its output space (DG1), surfacing what outputs share (DG2), and where they diverge (DG3). DG5 preserves raw-text access (\S3.5.5). DG4 and DG6 support comparing output sets across inputs and iterating on prompts and settings (\S3.5.3, \S3.5.6).}

\section{\tool{} System Design}
\label{sec:design}

Guided by the design goals above, we built \tool{} through an iterative design process. \edit{Early prototypes were developed in collaboration with lab members, whose feedback on layout, interaction, and information density shaped successive revisions (see below). We then refined the tool based on the formative study findings (\S\ref{sec:formative}), which grounded the design goals (DG1 to DG6). The resulting interface lets users compose prompts, inspect and filter outputs, and compare across prompts and model configurations in real time.} We evaluate key aspects of the tool quantitatively in three controlled user studies (\S\ref{sec:eval}).

\edit{We explored several alternative visual encodings during this process. Word trees~\cite{wattenberg2008word} (Appendix Fig.~\ref{fig:wordtree}) were easy to read and clearly showed which outputs shared prefixes (DG2), but are unable to surface the common pattern of LM outputs reconverging after diverging, sharing structure or phrasing later in the sentence (DG3). Color-coded span highlighting~\cite{10.1145/3613904.3642139} (Appendix Fig.~\ref{fig:highlights}) was useful for identifying shared segments (DG2), but still required users to read the full text of every output. This became burdensome at mesoscale sample sizes (DG1) and made it harder to see overall branching structure at a glance (DG3). A Time Curves-style encoding~\cite{Bach2016-timecurves} (Appendix Fig.~\ref{fig:timegraph}) offered another aggregate view of the output set, but did not preserve left-to-right readability of the underlying text, which we prioritized for inspecting LM generations. Based on these early lightweight tests and feedback, we moved toward the graph-based approach.}

\begin{figure*}[h]
  \centering
  \includegraphics[width=\linewidth]{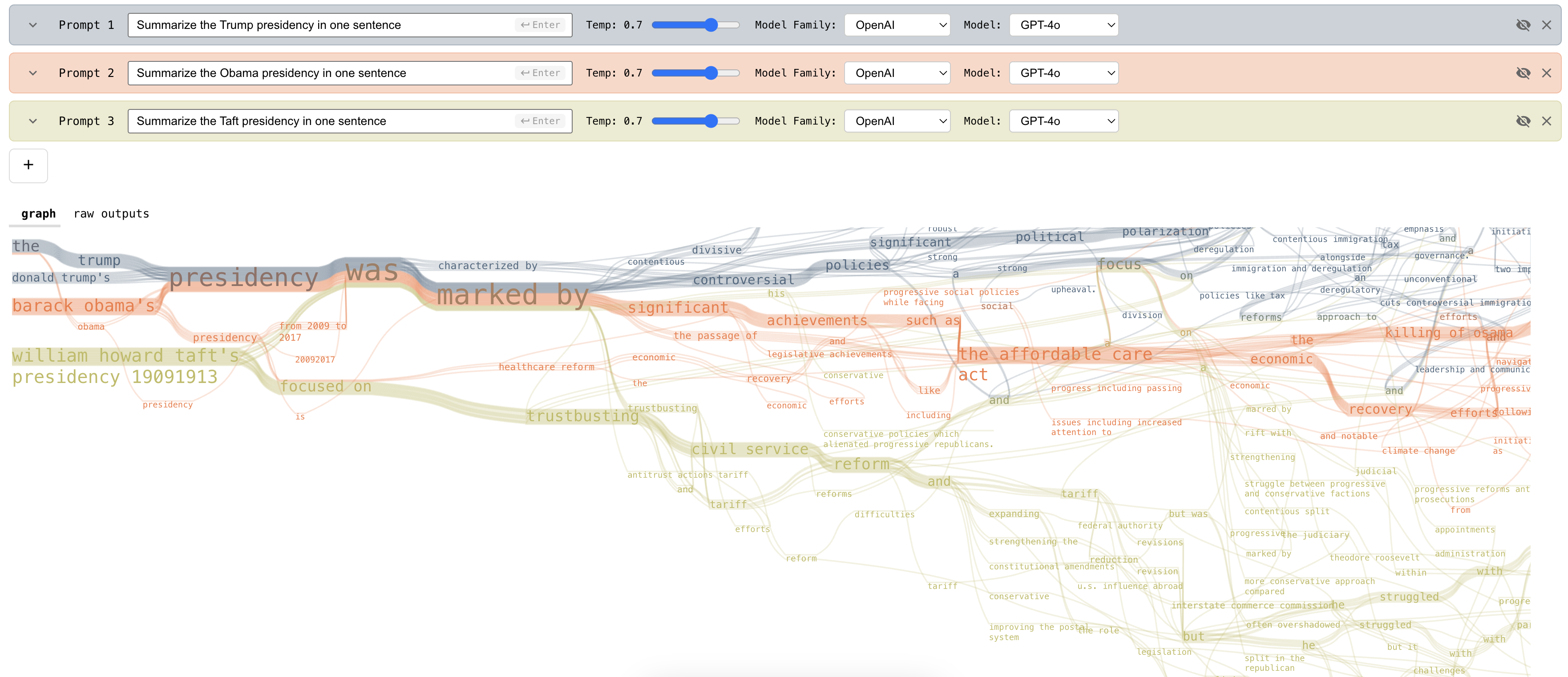}
  \caption{A comparison of the prompts ``summarize the [Trump/Obama/Taft] presidency in one sentence''. The nodes in the graph are colored by the prompt they correspond to. Obama and Trump's presidency descriptions have the same initial phrasing "presidency was marked by...", while the Taft presidency generations are more divergent.}
  \label{fig:presidents}
\end{figure*}

\subsection{Entering one or more prompts}
\edit{Fig.~\ref{fig:basic}}-A shows the prompt input box, which also includes controls for the temperature, number of generations, model family and model version (DG4, DG6). The interface defaults to a single input, but users can add prompts with the `+' button in \edit{Fig.~\ref{fig:basic}}-A.
The user can also adjust the number of generations to sample.

\subsection{Visualizing a merged token graph}
\tool{} shows a visual summary of the outputs as a graph of connected tokens, where each output is represented by a path through the graph \edit{(DG1--DG3)}, shown in \edit{Fig.~\ref{fig:basic}}

\subsubsection{Constructing the graph} \edit{To merge multiple model responses into one visualization, \tool{} processes all outputs through the following steps: first representing each output as a token chain, then merging those chains into a shared graph structure.}

\textbf{Tokenize outputs.} \tool{} tokenizes the outputs from all prompts according to the user-selected tokenization mode. The default is \textbf{space}, and the other options are \textbf{sentence} (splitting on periods, exclamations, or question marks followed by whitespace), or \textbf{phrase} (splitting on commas). Choosing space yields fine-grained word-level structure, and choosing sentence or phrase modes produce coarser, higher-level views. We did not have access to the model's internal tokenization scheme, so did not include this option.

\edit{\textbf{Create graph.} For each output, \tool{} adds a node for each token and a directed edge to the next token in that output, forming one linear chain per generation. At this point, the outputs remain separate chains rather than a single merged structure.}

\edit{\textbf{Merge tokens.} This step converts the set of output chains into the merged graph structure. \tool{} compares tokens \emph{across} outputs and collapses them into a single node when they match exactly or are semantically similar, using a heuristic inspired by Janicke et al.~\cite{janicke2014visualizations} and Gero et al.~\cite{10.1145/3613904.3642139}.} For each pair of tokens, we compute a semantic similarity score using an embedding model\footnote{\url{https://huggingface.co/Xenova/all-MiniLM-L6-v2}},  taking in the context of surrounding tokens. Tokens are merged when this score exceeds a user-adjustable threshold (default: 0.5), unless this would create a cycle. We also penalize merges between tokens that are far apart in the original sentence (see \S\ref{sec:alg-similarity} for the full algorithm).

\textbf{Collapse chains.} Chains of tokens without branches are collapsed into a single node to reduce visual clutter.

The resulting graph is a word lattice\cite{young1989token} over the tokenized outputs. We add directed links between each token and its successor whenever that adjacency appears in at least one generation. However, this can be misleading because some paths through the lattice do not correspond to any single sampled output (they ``mix'' prefixes and suffixes from different generations). 
To make it clear which sequences are real, we do not rely on the bare adjacency graph alone for rendering the graph. Instead, each generation is drawn as one continuous path that visits all of its tokens in order.

\subsubsection{Layout and rendering} The layout attempts to balance two competing goals: (1) preserve the left-to-right reading order of the original sentences so that tokens appearing sequentially in a generation also appear in sequence in the visualization, and (2) merge shared spans and expose the branching and merging structure of the graph. To this end, we use a D3 force simulation combining four forces, which is run until convergence. \edit{For force strengths, convergence criteria, and full implementation details, see our open-source code at \url{https://github.com/EmilyReif/llm-consistency-vis}.}
\begin{itemize}
    \item \edit{\textbf{Horizontal positioning.} A custom horizontal force updates each node's $x$-position based on its parents. First, root nodes anchor at a fixed left margin, and all other nodes are placed to the right of their parents according to parent width plus padding. When a node has multiple parents, a spread slider interpolates between the leftmost and rightmost parent-derived positions; one extreme gives a more linear, Sugiyama-style layered layout \cite{sugiyama1995graph}, while the other produces a more consolidated view.}
    \item \edit{\textbf{Vertical centering.} A vertical centering force pulls nodes toward the midpoint of the canvas, with strength proportional to each node's frequency count so that common tokens settle near the center and rarer tokens drift outward.}
    \item \edit{\textbf{Edges.} A constant link spring force connects consecutive tokens along each generation path. Because we instantiate one link object per generation traversing an edge, token pairs shared by more outputs experience stronger effective attraction.}
    \item \edit{\textbf{Collision.} A custom ellipse-based collision force prevents label overlap.}
\end{itemize}

Each token is colored by the weighted average of the colors of the sentences containing it. As described above, each edge represents a generation, and these are also colored corresponding to their prompt. The size of each node is proportional to the number of generations containing it, following other similar tools\cite{wattenberg2008word, Brockett2015-tq}.

\subsection{Interacting with the graph}
\edit{\textbf{Simplification (DG1, DG2)}}
We found a tension in the formative study on what level of summarization was appropriate. Participants want an unfiltered view of the outputs (that is, they want to see all outputs in the graph). However, this could be overwhelming, so they wanted a way to simplify the visualization as well to see the main modes. To avoid sacrificing either, we provided two interactions for simplifying the graph. The \textbf{hide longtail} slider shows the most frequent words and phrases by giving a lower opacity to nodes that appeared in few outputs. While this makes underlying outputs less legible (since sections of the text might not be shown), it highlights common words and phrases, and is akin to methods like word clouds~\cite{6758829}. The \textbf{merge similar} slider adjusts the semantic similarity threshold used during graph construction (\S\ref{sec:alg-similarity}). Lowering it merges more tokens, producing a more consolidated graph in which near-synonymous spans are collapsed.

\edit{\textbf{Filtering (DG2, DG5)}}
Users can also select a node to filter the graph to only outputs containing it. When they do this, any examples \textbf{not} containing that node are visually de-emphasized (blurred and lowered opacity), and the graph layout algorithm is recomputed. This follows a focus / context approach \cite{Furnas1986TheFV}: the selected subgraph is foregrounded while the surrounding structure remains visible, letting users drill into a specific region without losing a sense of where it fits in the overall distribution. See \edit{Fig.~\ref{fig:compare_filter}} for examples; \edit{Fig.~\ref{fig:view_subset}} shows the node-selection interaction.

\textbf{Switching between graph and plain text (DG5)} Users repeatedly stressed that summary views obscure important details. Thus, the interface keeps raw outputs easily accessible. Because the graph sacrifices legibility of individual examples for summarization, the plain-text responses are also shown on demand, with optional filtering. The list and graph are cross-linked: selecting a listed output highlights its corresponding path through the graph, and graph interactions (including node-based filtering in the previous subsection) propagate to the list so that matching completions are easy to spot while non-matching ones stay visible but de-emphasized.

\subsection{Comparison mode (DG4)} Comparing distributions was a core need described by participants (DG4), both for prompt iteration and other types of prompt comparison. The user clicks the `+' button to add another prompt, which they can configure as before. The resulting graphs are color coded by prompt, and can be viewed as a single merged visualization, or as separate side by side visualizations.

\section{Evaluation: Comparing Graph and List Views}
\label{sec:eval}

\edit{Our formative study with LM practitioners (\S\ref{sec:formative}) showed that they already reason distributionally over many outputs per prompt, but often lack tools beyond unstructured lists. We designed three studies to span the most common workflows formative participants described, covering prompt iteration, debugging common failure modes, consistency analysis, and model-behavior auditing. Prompt iteration was the most-mentioned workflow (\S3.5.3), motivating both the diversity comparison (Study~1) and two-distribution comparison (Study~3) tasks, which compare output sets across two similar prompts. Failure-mode debugging and consistency analysis often reduce to a diversity judgment, since participants described failures such as mode collapse or runaway variation (\S3.5.3); the diversity study targets these directly. The single-distribution comprehension task (Study~2) was grounded in concrete formative experiences such as P1's observation that her boy-and-pet stories repeatedly featured ``a boy named Joe and a dog named Fido'' (\S3.5.4), motivating questions on the frequency of named entities and recurring phrases.}
\tool{} includes several features (graph simplification and filtering) from interview feedback, but our evaluation focuses on the core question: for which distributional reasoning tasks does a merged graph summary help, and for which does a plain list suffice?

\edit{We used crowdsourced participants role-playing prompt iteration rather than formative-study experts to support within-subjects comparison with adequate power. This study was approved by the IRB of the University of Washington (No. STUDY00024291); participants provided informed consent via the platform.}

\textbf{Conditions.} The only manipulation was how the model outputs were presented. In the \textbf{graph} condition, participants used the merged-token graph visualization described in \S\ref{sec:design}. In the \textbf{list} condition, they saw those same outputs as a scrollable plain-text list, with batches separated by prompt whenever two prompts were in play. The list presentation is intentionally simple (no merged structure); formative participants nonetheless reported that they often assess models by reading many raw outputs, so this condition mirrors a common practice.

\textbf{Procedure.} Each study used a within-subjects design: \edit{each participant used both interfaces, with a different dataset assigned to each interface}, with interface assignment, dataset assignment, and ordering counterbalanced. A final comparative survey collected subjective preferences. Participants role-played as game designers working on prompts for a fantasy game to develop both in-game monsters and places. Outputs were pre-cached from two datasets: \textit{monsters} (creature descriptions) and \textit{places} (location descriptions). We chose these tasks for their open-endedness, desirable diversity, and accessibility to non-LM experts. \edit{See Figs.~\ref{fig:user-study-diversity} and~\ref{fig:user-study-single} for screenshots}.

All three studies shared the same subjective measures. After each task, participants rated both interfaces on 7-point Likert scales assessing understanding of model behavior (diversity, typical outputs, rare outputs, recurring patterns), workload and effort (NASA-TLX), and usability and satisfaction. They also completed direct comparison questions on a 7-point forced-choice scale (1 = graph, 7 = list) asking which interface they preferred overall and on specific dimensions (e.g., ease of use, confidence). Full question wording is included in the Post-Task Survey Protocol (\S\ref{sec:post-task-survey}).

\subsection{Studies and tasks}

\subsubsection{Diversity comparison}
\textbf{Research question.} Does the graph help users determine the diversity difference between two output distributions?
\edit{Formative study participants emphasized various forms of diversity being important in many cases, including iterating on prompts (\S3.5.3), and understanding failure modes.}

\textbf{Task.} $N{=}47$ participants saw two side-by-side sets of outputs from the
same prompt but generated at different temperature settings (differing
by 0.1 on a 0.1 to 1.0 scale). Their task was to judge which set had higher diversity. Each interface block was timed at 3~minutes, during which participants answered as many pairwise comparisons as
possible.

\textbf{Measures.} Each comparison had an objective ground truth (the higher-temperature set was deemed more diverse). We measured accuracy (fraction correct) and time per question for each interface, subtracting loading and rendering time. Participants also completed the shared subjective measures.

\subsubsection{Single distribution comprehension}
\textbf{Research question.} Does the graph visualization improve users' understanding of a single prompt's output distribution?
\edit{Formative study participants described distributional sensemaking in qualitative terms: identifying typical outputs, recurring patterns, and outliers without agreed-upon metrics (\S3.5.4). We operationalize these judgments as structured questions over a fixed sample of $k{=}20$ outputs.}

\textbf{Task.} $N{=}44$ participants viewed $k{=}20$ outputs from one prompt and answered a structured questionnaire about the distribution. All items were scored against ground truth derived from those same 20 outputs. Most items have a single correct choice (e.g., which proper name appears most often; which output is an out-of-domain implausibility; which displayed output was actually sampled). Theme and frequent-phrase items use a short list of candidate paraphrases and are scored with flexible string matching. Examples are shown in \S\ref{sec:example-task-questions}.

\textbf{Measures.} We report accuracy: exact match where appropriate, partial credit on percentage-bucket distance, and the flexible matching rules in \S\ref{sec:example-task-questions} for theme and phrase items. Participants also completed the shared subjective measures.

\subsubsection{Two-distribution comparison}
\textbf{Research question.} Does the graph interface help users identify specific differences in the output distributions of two prompts?
\edit{When comparing prompt variants, formative study participants asked if observed differences were meaningful or noise (\S3.5.3). This study tests attribution and relative-frequency judgments across two hidden prompts.}

\textbf{Task.} $N{=}40$ participants viewed $k{=}20$ outputs from each of two prompts (displayed as Prompt~1 and Prompt~2, with prompt text hidden) and answered a structured questionnaire about them. All items were scored against ground truth derived from those same 20 outputs: relative frequency (which prompt more often mentions a named entity or phrase), attribution (which prompt likely produced a shown sentence), and discrimination (outputs plausible under one, both, or neither prompt). Examples are shown in \S\ref{sec:example-task-questions}.

\textbf{Measures.} We report accuracy against ground truth from the displayed outputs, using the same partial-credit rule for percentage bins where applicable (\S\ref{sec:example-task-questions}). Participants also completed the shared subjective measures.

\subsection{Results}

\begin{figure}[h]
  \centering
  \includegraphics[width=\linewidth]{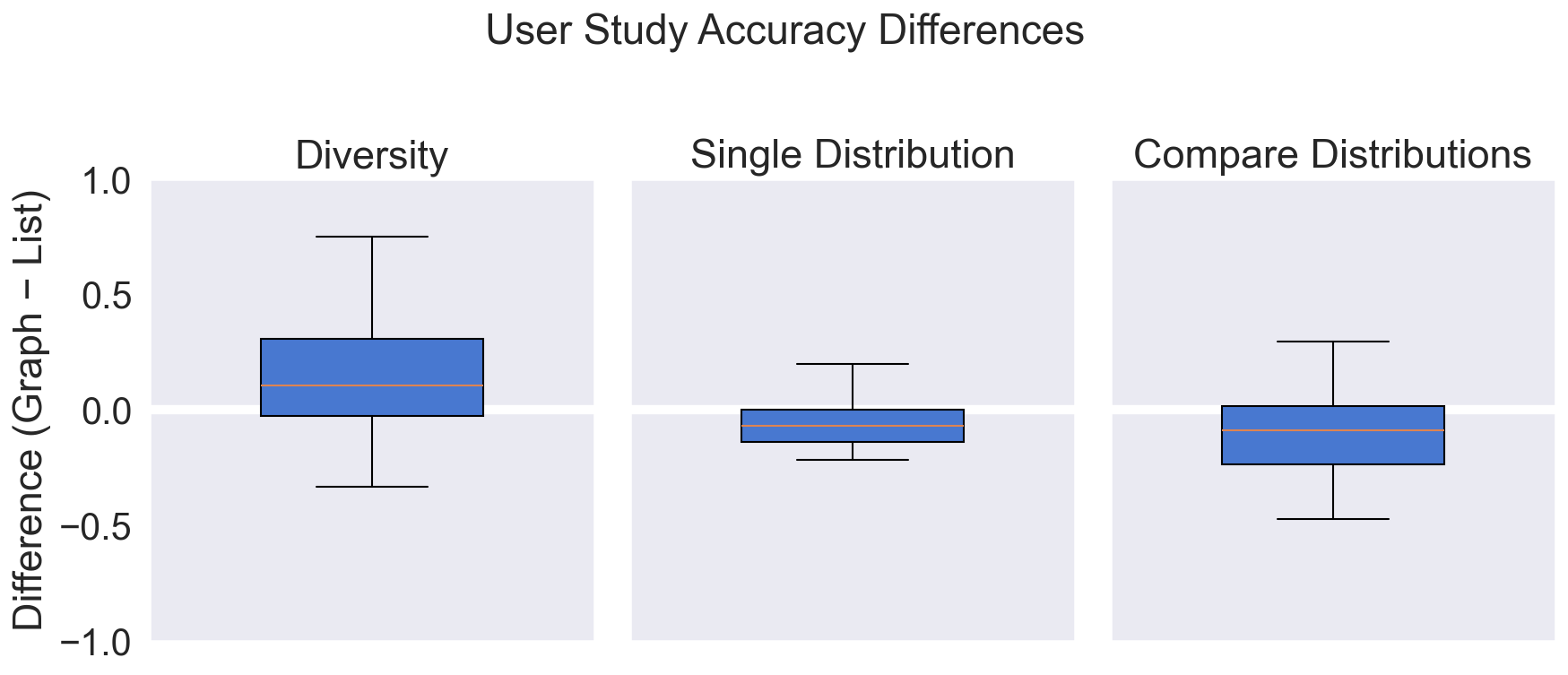}
  \caption{Per-participant difference (Graph $-$ List) in accuracy for all three studies. Diversity ($n{=}36$): graph yielded higher accuracy, $p{=}0.012$. Single distribution ($n{=}26$): list yielded higher accuracy, $p{=}0.009$. Comparison ($n{=}40$): list yielded higher accuracy, $p{=}0.002$.}
  \label{fig:combined_diff}
\end{figure}

We analyzed accuracy and time using Wilcoxon signed-rank tests~\cite{wilcoxon1945} on the per-participant difference (Graph $-$ List), including only participants who completed both conditions (some ran out of time on one interface and were excluded from paired analyses: 11 of 47 in Study~1, 18 of 44 in Study~2, and none in Study~3). \edit{Fig.~\ref{fig:combined_diff}} shows the accuracy differences across all three studies.

\begin{figure}[h]
  \centering
  \includegraphics[width=\linewidth]{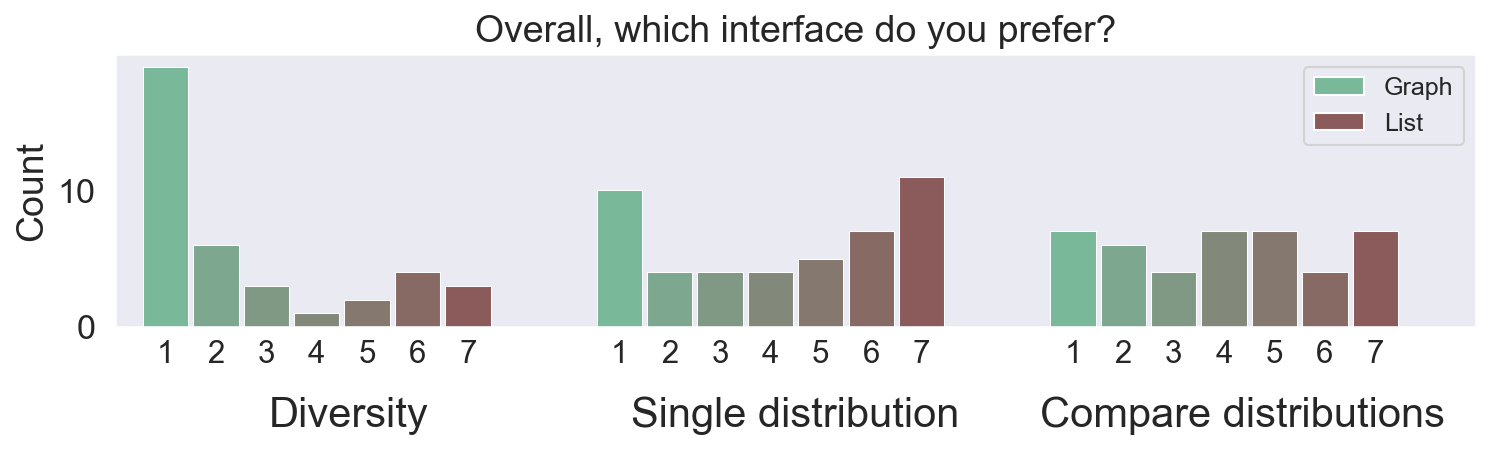}
  \caption{Overall interface preference (1 $=$ graph, 7 $=$ list) by study. For diversity, participants strongly favored the graph. For single distribution, preferences were polarized. For the comparison task, preferences were more evenly spread.}
  \label{fig:direct-comparison-preference}
\end{figure}
\textbf{Summary.} For the diversity task, the graph yielded higher accuracy, faster completion, and stronger preference (\edit{Fig.~\ref{fig:direct-comparison-preference}}), but for single-distribution and comparison tasks, the list yielded higher accuracy (time differences were not significant). Preferences varied by task: strongly pro-graph for diversity; polarized for single distribution (participants strongly favored either graph or list); and more uniform for comparison.

\subsubsection{Diversity comparison}
Users were significantly more accurate at evaluating the relative diversity of text distributions with the graph than with the list (mean difference $=$ 0.12, 95\% CI $[0.03, 0.21]$, Wilcoxon $p{=}0.012$, $n{=}36$). Participants were also faster with the graph (mean $=$ $-10.9$\,s, 95\% CI $[-21.8, 0.1]$\,s, $p{=}0.052$). Participants strongly favored the graph on all direct comparison dimensions and found the list more overwhelming (\edit{Fig.~\ref{fig:direct-comparison-preference}}; \edit{Fig.~\ref{fig:direct-comparison-all}}).

\subsubsection{Single distribution comprehension}
For the task of comprehending a single distribution, the list significantly outperformed the graph in accuracy (mean difference $=$ $-0.057$, 95\% CI $[-0.098, -0.015]$, Wilcoxon $p{=}0.009$, $n{=}26$ participants with both conditions). Participants spent about 1.0\,min more on the graph on average (95\% CI $[-0.39, 2.41]$\,min), but the time difference was not statistically significant ($p{=}0.15$). Preferences were polarized, with responses clustered toward both ends of the 7-point overall-preference scale (\edit{Fig.~\ref{fig:direct-comparison-preference}}). Users' preferences weakly correlate with accuracy (r=0.23, p=0.27) and time (r=0.38, p=0.06), but these relationships are not statistically significant. 


\subsubsection{Two-distribution comparison}
The list also significantly outperformed the graph in accuracy for the task of comparing distributions (mean difference $=$ $-0.10$, 95\% CI $[-0.16, -0.04]$, Wilcoxon $p{=}0.002$, $n{=}40$). A breakdown by question type is in \edit{Fig.~\ref{fig:compare-accuracy-by-qtype}}. Participants spent about 58\,seconds more on the graph on average (95\% CI $[-15.6, 132.0]$\,s), but the time difference was not statistically significant ($p{=}0.12$). Unlike single distribution, preferences were more evenly spread across the scale, rather than being polarized toward one interface or the other (\edit{Fig.~\ref{fig:direct-comparison-preference}}).

\begin{figure}[h]
  \centering
  \includegraphics[width=\linewidth]{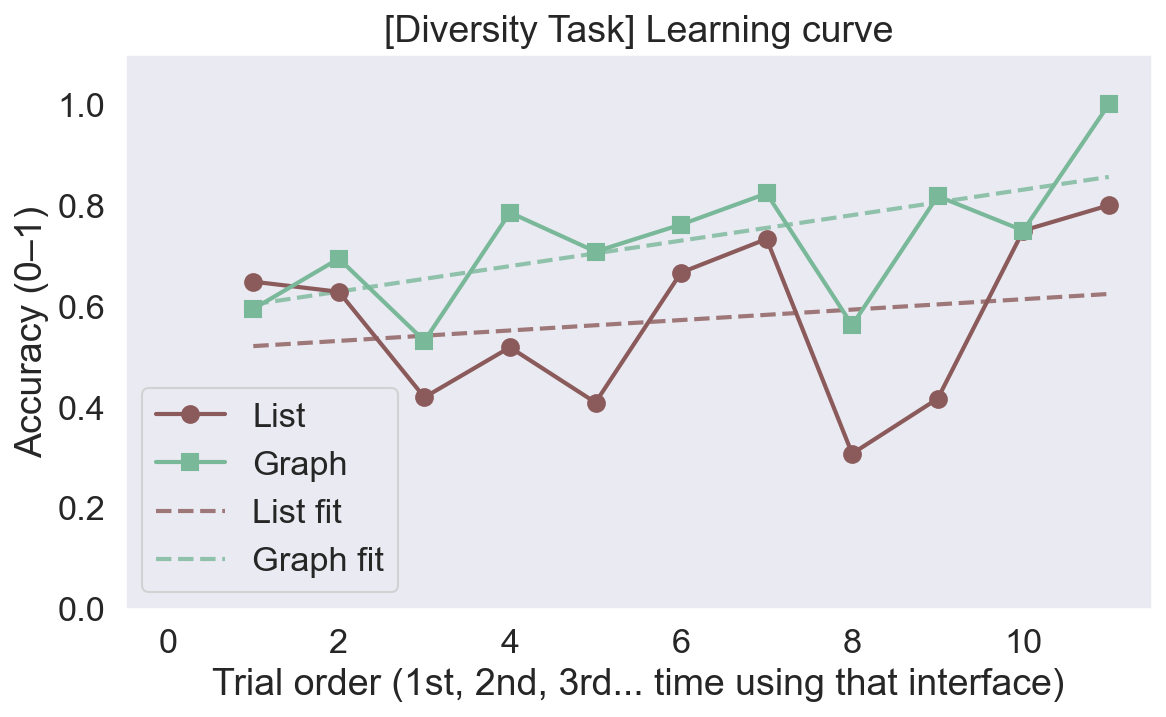}
  \caption{Learning curve for the diversity task. Proportion correct by trial order (1st, 2nd, 3rd\ldots{} time using that interface). The graph showed a slightly steeper learning curve than the list.}
  \label{fig:diversity_learning_curve}
\end{figure}

\subsubsection{Post-task subjective measures}

\textbf{Direct comparison.} For the forced-choice questions (1 $=$ graph, 7 $=$ list), preferences varied by task (\edit{Fig.~\ref{fig:direct-comparison-preference}}). On ``which interface better supported understanding the range and distribution of outputs,'' participants favored the graph for all three tasks. The remaining direct comparison figures are in \edit{Fig.~\ref{fig:direct-comparison-all}}. For the independently rated Likert scales (one set per interface), Wilcoxon signed-rank tests showed no significant differences on any question ($p > 0.05$ for all).

\edit{\textbf{Qualitative feedback on interfaces.} Participants favored the graph for patterns (``spotting `mode collapse,''' ``bird's-eye view of the AI's `brain,''' ``thick lines\ldots at a glance'') but found it ``visually dense'' and ``hard to follow''; lists were ``easy to scan and read'' but required them to ``scroll a lot and remember what you read 20 lines ago,'' and made it ``hard to see the `average' behavior.'' Most also viewed the combined interface positively (80\% positive).}

\FloatBarrier
\section{Discussion}

\begin{figure}[htbp]
  \centering
  \includegraphics[width=\linewidth]{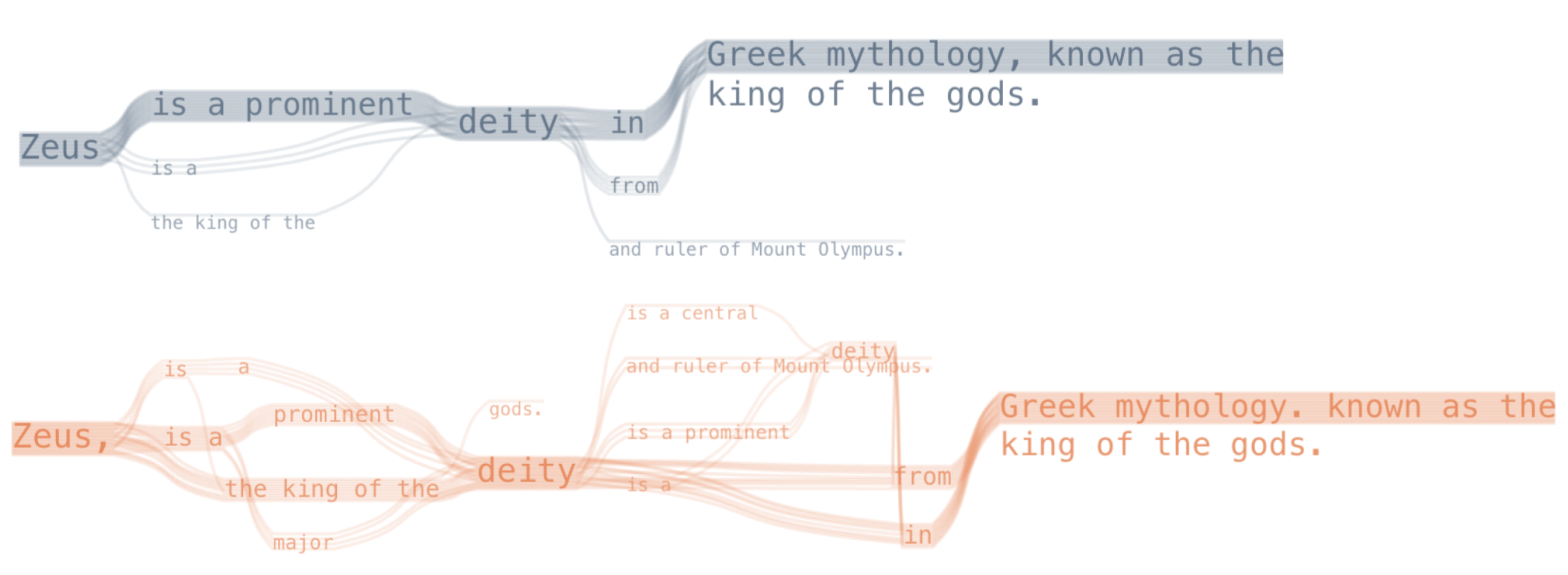}
  \caption{Outputs from the open-ended prompt ``name a Greek god or goddess''. Across temperatures from 0.1 (blue) to 0.9 (orange), the model consistently returns Zeus, varying only in phrasing.}
  \label{fig:zeus}
\end{figure}

\edit{Overall, our findings suggest that hybrid graph-and-list interfaces are a promising direction for distributional reasoning. Read together with the formative study (\S\ref{sec:formative}), the evaluation points to a task-dependent split: the graph helped with high-level diversity judgments that formative participants described as important but hard to make over long lists of raw outputs (Study 1), while list inspection remained stronger for detail-oriented questions (specific entities, attribution, verification) that mirror formative participants' default notebook workflows (Studies 2 and 3). This aligns with prompt-iteration workflows where practitioners first assess overall variation, then audit individual generations. Qualitative feedback reinforced this tradeoff: graphs surface patterns and frequency at a glance but can feel visually dense; lists support detailed reading but overwhelm at scale.}

\textbf{Text distributions vary widely, and visualization effectiveness depends on their structure.}
A key takeaway from both the formative study and evaluation is that the usefulness of any visualization depends heavily on the structure of the underlying text distribution. LM outputs can differ in many ways: they may be tightly aligned or highly divergent, short or long, dominated by a few repeated templates or widely varied in phrasing. These properties strongly influence whether structural visualization techniques are helpful.

\begin{figure*}[t]
  \centering
  \begin{subfigure}[t]{\linewidth}
    \includegraphics[width=\linewidth]{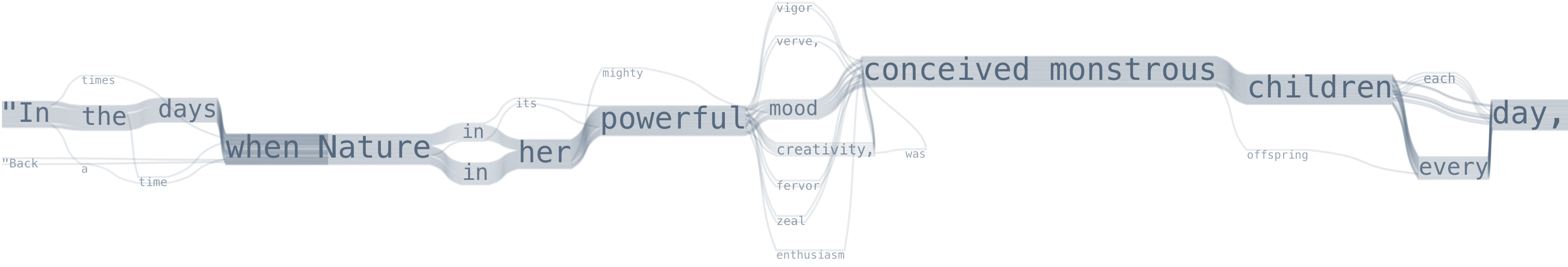}
    \caption{\textbf{Branching and reconverging:} Translation outputs are often temporally aligned, with small points of divergence. The graph of English translations of La Géante, a French poem, reveals where translations branch and reconverge.}
    \label{fig:translation}
  \end{subfigure}
  \vspace{\smallskipamount}
  \begin{subfigure}[t]{\linewidth}
    \includegraphics[width=\linewidth]{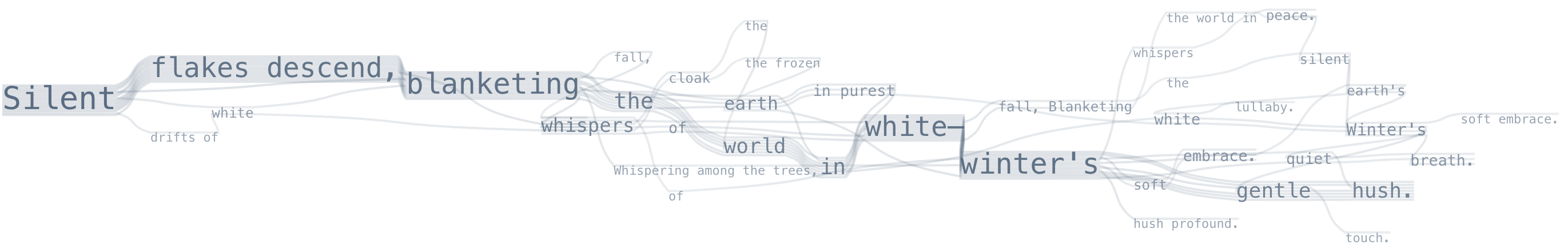}
    \caption{\textbf{Surfacing patterns:} Creative prompts such as haikus often share repeated framing phrases or structural templates. The graph surfaces these patterns.}
    \label{fig:haiku}
  \end{subfigure}
  \vspace{\smallskipamount}
  \begin{subfigure}[t]{\linewidth}
    \includegraphics[width=\linewidth]{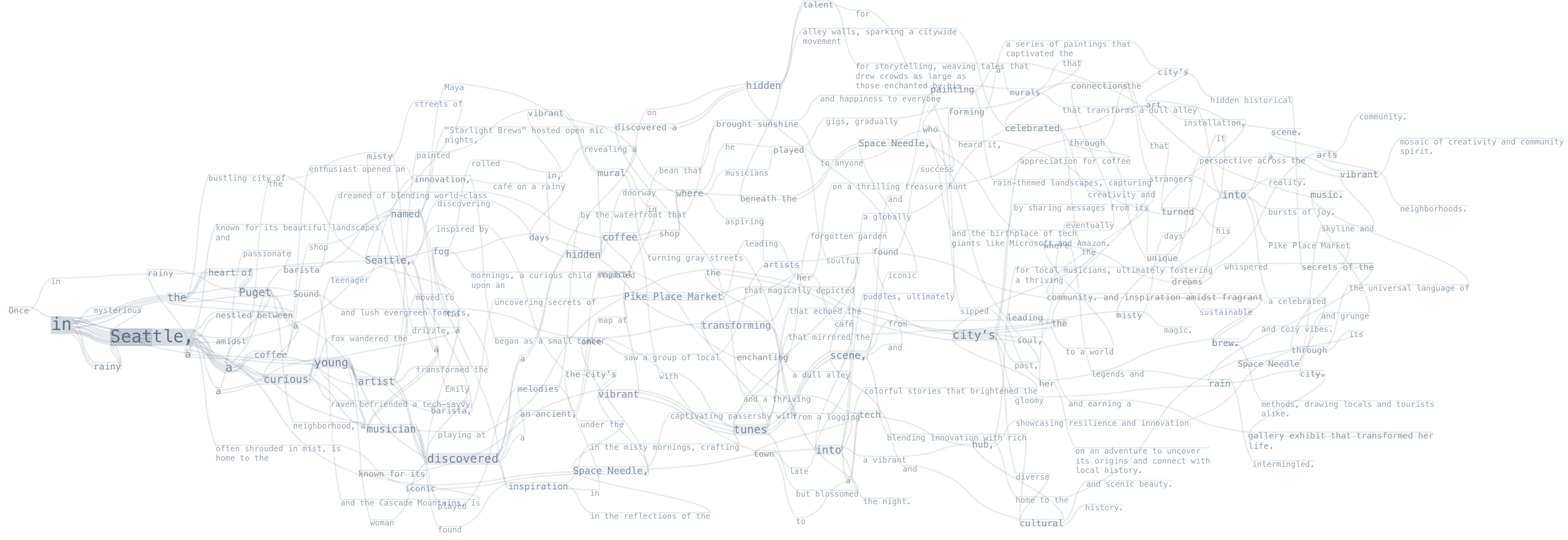}
    \caption{\textbf{Divergent and hard to read:} Outputs from the prompt ``tell me a story about Seattle'' with temp $=$ 1. The graph visualization is unreadable for very highly divergent outputs, except to tell you that they are indeed divergent.}
    \label{fig:hairball}
  \end{subfigure}
  \caption{Examples of how graph visualizations surface distributional structure: (a) temporally aligned translation outputs, (b) structural templates in creative writing, and (c) highly divergent outputs where the graph becomes unreadable.}
\end{figure*}

Visual encodings that expose shared structure are particularly effective when outputs are temporally or syntactically aligned. For example, tasks like translation or style rewriting tend to produce outputs with similar lengths and structure but small points of divergence. In these settings, structural visualizations make it easy to see where outputs branch and reconverge (\edit{Fig.~\ref{fig:translation}}).

Similarly, distributions that contain a few strong modes can benefit from structural summaries. For example, outputs to ``name a Greek god or goddess'' repeatedly return Zeus with only superficial variation across temperature settings (\edit{Fig.~\ref{fig:zeus}}). \edit{Appendix Fig.~\ref{fig:random_numbers}} also shows a graph of ``Give me a random number between 1 and 100'' for comparison, highlighting how models are also poor random number generators  \cite{hopkins2023can}. We also observed this pattern in more open-ended tasks. Creative prompts such as poems or stories also often share repeated framing phrases or structural templates (e.g., similar openings or narrative patterns). When these patterns exist, visualizing shared structure helps reveal them quickly (\edit{Fig.~\ref{fig:haiku}}).

However, structural encodings become less effective when outputs are \textbf{long, highly heterogeneous, or only semantically similar while differing token-by-token}. In these cases, branching can become extremely dense, producing visually complex layouts without much compression benefit (\edit{Fig.~\ref{fig:hairball}}). Outputs that diverge early and rarely reconverge are particularly difficult to summarize visually.

Two additional simple properties shape this design space. The first is \textbf{output length}. Short responses often reveal structure clearly, whereas long-form generations are more difficult to interpret. Some filtering or summarization may help highlight key areas of overlap without getting overwhelmed by the sheer amount of text. 
The second is the \textbf{number of outputs}. The ``mesoscale'' range of tens to hundreds of generations (defined by Gero et al.~\cite{10.1145/3613904.3642139} and similar to what practitioners reported inspecting in the formative study) appears well suited to visual analysis. Smaller sets can be read directly. Pushing to much larger sets, or relying on finite samples for claims about the full output distribution, runs into limits of our encoding and of sampling itself; we discuss these in \S\ref{sec:limitations-future}.

Taken together, these observations suggest that understanding which properties of text distributions make structural visualization useful is an important direction for future work. A central question we can explore is: what aspects of a text distribution should a visualization surface, and which encodings are best suited to doing so? 


\textbf{Opportunities for visualization in generative AI workflows.} As LMs become components in larger systems, practitioners must reason about collections of related but non-identical outputs with no clear ground truth. Current interfaces mostly show one completion at a time, leaving users to infer distributional behavior implicitly. Visualization can make patterns, modes, and divergences visible and move interaction beyond ``prompting in the dark.'' Tools like \tool{}, along with clustering, summaries, and other hybrid encodings, are one step toward more informed generative-AI workflows.

\section{Limitations / Future Work / Conclusion}
\label{sec:limitations-future}

\edit{\textbf{Lab evaluation.}} Our evaluation highlighted a hybrid need: a graph for high-level patterns and a list for detailed reading, with different interfaces better suited to different tasks. While our tool addresses this to some degree, we do not evaluate that feature as part of this study, as we were more focused on different strengths and weaknesses of different encoding strategies. \edit{As a general limitation, our crowdsourced tasks test these options under controlled conditions rather than with formative-study experts in situ; validating these patterns in real workflows remains important future work.}

\edit{\textbf{Sample size and visual legibility.}} For many statistical summaries (e.g., a histogram), drawing more samples yields a smoother, more stable aggregate. Our encoding chooses a different tradeoff: each generation must appear as a complete path through the merged graph, so for some types of distributions, additional samples add more overlapping paths and the display becomes hairball-like, rather than converging to a simple aggregate. Future work could relax that invariant (e.g., by aggregating or subsampling paths) and evaluate when such designs improve legibility and when they risk misleading users about the underlying distribution. Relatedly, \tool{} always visualizes a finite sample from the model's output distribution, not the distribution itself. A single generation is usually a weak basis for claims about variability or ``typical'' behavior, but even tens, hundreds, or more outputs can still miss rare modes or misrepresent tail behavior depending on the prompt and decoding settings. Future interfaces could also incorporate model internals beyond discrete output samples, such as token-level probabilities.

\edit{\textbf{Qualitative and quantitative diversity.}} Another direction is to better connect qualitative notions of diversity with quantitative measures. Users often begin with informal, qualitative judgments and then attempt to operationalize them. Future systems could support this process more directly, enabling rapid translation from qualitative insight to task-specific metrics.

\edit{\textbf{Extensions beyond text.} \tool{}'s token graph, semantic merge, and left-to-right layout are text-specific, but the underlying pattern (each sample as a path through a merged structure, with raw outputs preserved) may apply more broadly. We focus on LM text here, but modern AI workflows also involve tables, tool calls, and images, which we leave to future work. In exploratory lab work with collaborators, we visualized machine transcriptions of different pianists playing the same piece in \tool{} by feeding the transcription tokens directly into the graph. These are not LM outputs, but they share temporally aligned sequential structure with localized variation across performances. The graph was useful for comparing how performances differed, but a visualization anchored to the musical score itself, leveraging the piece's overall structure rather than transcription tokens alone, would likely be more appropriate.}

Overall, treating LM behavior as a distribution rather than as a single answer significantly changes the interface's design requirements. We discovered design implications through a formative study, proposed a graph-based multi-distributional LM output visualization as \tool{}, and conducted controlled studies that show that structural summaries can complement, but not replace, reading raw outputs. We hope this line of work encourages more tools and evaluations that make distributional structure a first-class part of how people prompt, debug, and judge language models.

\acknowledgments{%
We thank members of the Interactive Data Lab, especially Madeleine Grunde-McLaughlin, Hyeok Kim, and Ameya Patil, as well as Noah’s ARK Lab, Adam Pearce, Martin Wattenberg, and Fernanda Viégas for their valuable feedback and discussions. Emily Reif is supported by the Amazon AI PhD Fellowship. This material is based upon work supported by the National Science Foundation under Award No. 2413244.
}
\clearpage

\bibliographystyle{abbrv-doi-hyperref}
\bibliography{sample-base}

@String{Computing = "Computing" }

@String{Computer = "{IEEE} Computer" }

@article{sugiyama1995graph,
  title={Graph drawing by the magnetic spring model},
  author={Sugiyama, Kozo and Misue, Kazuo},
  journal={Journal of Visual Languages \& Computing},
  volume={6},
  number={3},
  pages={217--231},
  year={1995},
  publisher={Elsevier},
  doi={10.1006/jvlc.1995.1013}
}

@inproceedings{collins2007visualization,
  title={Visualization of uncertainty in lattices to support decision-making},
  author={Collins, Christopher and Carpendale, Sheelagh and Penn, Gerald},
  booktitle={Proceedings of the Eurographics/IEEE-VGTC Symposium on Visualization},
  pages={51--58},
  year={2007},
  doi={10.2312/VisSym/EuroVis07/051-058}
}

@article{riehmann2012wordgraph,
  title={{WORDGRAPH}: Keyword-in-context visualization for {NETSPEAK}'s wildcard search},
  author={Riehmann, Patrick and Gruendl, Henning and Potthast, Martin and Trenkmann, Martin and Stein, Benno and Froehlich, Bernd},
  journal={IEEE Transactions on Visualization and Computer Graphics},
  volume={18},
  number={9},
  pages={1411--1423},
  year={2012},
  publisher={IEEE},
  doi={10.1109/TVCG.2012.96}
}

@inproceedings{janicke2014visualizations,
  title={Visualizations for text re-use},
  author={J{\"a}nicke, Stefan and Ge{\ss}ner, Annette and B{\"u}chler, Marco and Scheuermann, Gerik},
  booktitle={Proceedings of the International Conference on Information Visualization Theory and Applications (IVAPP)},
  pages={59--70},
  year={2014},
  publisher={SciTePress},
  doi={10.5220/0004692500590070}
}

@article{hu2016visualizing,
  title={Visualizing social media content with {SentenTree}},
  author={Hu, Mengdie and Wongsuphasawat, Krist and Stasko, John},
  journal={IEEE Transactions on Visualization and Computer Graphics},
  volume={23},
  number={1},
  pages={621--630},
  year={2017},
  publisher={IEEE},
  doi={10.1109/TVCG.2016.2598590}
}

@inproceedings{10.1145/3613904.3642139,
author = {Gero, Katy Ilonka and Swoopes, Chelse and Gu, Ziwei and Kummerfeld, Jonathan K. and Glassman, Elena L.},
title = {Supporting Sensemaking of Large Language Model Outputs at Scale},
year = {2024},
isbn = {9798400703300},
publisher = {Association for Computing Machinery},
address = {New York, NY, USA},
url = {https://doi.org/10.1145/3613904.3642139},
doi = {10.1145/3613904.3642139},
booktitle = {Proceedings of the 2024 CHI Conference on Human Factors in Computing Systems},
articleno = {838},
numpages = {21},
location = {Honolulu, HI, USA},
series = {CHI '24}
}

@INPROCEEDINGS{Gu2025-bf,
  title     = "{AbstractExplorer}: Leveraging structure-mapping theory to
               enhance comparative close reading at scale",
  author    = "Gu, Ziwei and Zhou, Joyce and Lei, Ning-Er (nina) and Kummerfeld,
               Jonathan K and Jasim, Mahmood and Mahyar, Narges and Glassman,
               Elena L",
  booktitle = "Proceedings of the 38th Annual ACM Symposium on User Interface
               Software and Technology",
  publisher = "ACM",
  address   = "New York, NY, USA",
  pages     = "1--25",
  month     =  sep,
  year      =  2025,
  language  = "en",
  doi       = {10.1145/3746059.3747773}
}

@misc{Zhang2025-sw,
  title         = {{NoveltyBench}: Evaluating Language Models for Humanlike Diversity},
  author        = {Zhang, Yiming and Diddee, Harshita and Holm, Susan and Liu,
                   Hanchen and Liu, Xinyue and Samuel, Vinay and Wang, Barry and
                   Ippolito, Daphne},
  year          = {2025},
  eprint        = {2504.05228},
  archivePrefix = {arXiv},
  primaryClass  = {cs.CL},
  doi           = {10.48550/arXiv.2504.05228},
  url           = {https://arxiv.org/abs/2504.05228}
}

@INPROCEEDINGS{He2025-ij,
  title     = "Prompting in the dark: Assessing human performance in prompt
               engineering for data labeling when gold labels are absent",
  author    = "He, Zeyu and Naphade, Saniya and Huang, Ting-Hao Kenneth",
  booktitle = "Proceedings of the 2025 CHI Conference on Human Factors in
               Computing Systems",
  publisher = "ACM",
  address   = "New York, NY, USA",
  pages     = "1--33",
  month     =  apr,
  year      =  2025,
  language  = "en",
  doi       = {10.1145/3706598.3714319}
}

@INPROCEEDINGS{Li2025-dl,
  title     = "{ReasonGraph}: Visualization of reasoning methods and extended
               inference paths",
  author    = "Li, Zongqian and Shareghi, Ehsan and Collier, Nigel",
  booktitle = "Proceedings of the 63rd Annual Meeting of the Association for
               Computational Linguistics (Volume 3: System Demonstrations)",
  publisher = "Association for Computational Linguistics",
  pages     = "140--147",
  year      = {2025},
  doi       = {10.18653/v1/2025.acl-demo.14}
}

@misc{Zhang2024-lx,
  title        = {Forcing Diffuse Distributions out of Language Models},
  author       = {Zhang, Yiming and Schwarzschild, Avi and Carlini, Nicholas and
                  Kolter, J. Zico and Ippolito, Daphne},
  year         = {2024},
  note         = {First Conference on Language Modeling (COLM 2024)},
  eprint       = {2404.10859},
  archivePrefix = {arXiv},
  primaryClass = {cs.CL},
  doi          = {10.48550/arXiv.2404.10859},
  url          = {https://arxiv.org/abs/2404.10859}
}

@INPROCEEDINGS{Arawjo2024-rw,
  title     = "{ChainForge}: A visual toolkit for prompt engineering and {LLM}
               hypothesis testing",
  author    = "Arawjo, Ian and Swoopes, Chelse and Vaithilingam, Priyan and
               Wattenberg, Martin and Glassman, Elena L",
  booktitle = "Proceedings of the CHI Conference on Human Factors in Computing
               Systems",
  publisher = "ACM",
  address   = "New York, NY, USA",
  pages     = "1--18",
  month     =  may,
  year      =  2024,
  language  = "en",
  doi       = {10.1145/3613904.3642016}
}

@INPROCEEDINGS{Cheng2024-xs,
  title     = "{RELIC}: Investigating large language model responses using
               self-consistency",
  author    = "Cheng, Furui and Zouhar, Vilém and Arora, Simran and Sachan,
               Mrinmaya and Strobelt, Hendrik and El-Assady, Mennatallah",
  booktitle = "Proceedings of the CHI Conference on Human Factors in Computing
               Systems",
  publisher = "ACM",
  address   = "New York, NY, USA",
  pages     = "1--18",
  month     =  may,
  year      =  2024,
  language  = "en",
  doi       = {10.1145/3613904.3641904}
}

@INPROCEEDINGS{Kahng2024-mm,
  title     = "{LLM} comparator: Visual analytics for side-by-side evaluation of
               large language models",
  author    = "Kahng, Minsuk and Tenney, Ian and Pushkarna, Mahima and Liu,
               Michael Xieyang and Wexler, James and Reif, Emily and
               Kallarackal, Krystal and Chang, Minsuk and Terry, Michael and
               Dixon, Lucas",
  booktitle = "Extended Abstracts of the CHI Conference on Human Factors in
               Computing Systems",
  publisher = "ACM",
  address   = "New York, NY, USA",
  pages     = "1--7",
  month     =  may,
  year      =  2024,
  language  = "en",
  doi       = {10.1145/3613905.3650755}
}

@INPROCEEDINGS{Hamilton2024-sl,
  title     = "Detecting Mode Collapse in Language Models via Narration",
  author    = "Hamilton, Sil",
  booktitle = "Proceedings of the First edition of the Workshop on the Scaling
               Behavior of Large Language Models ({SCALE-LLM} 2024)",
  address   = "St. Julian's, Malta",
  publisher = "Association for Computational Linguistics",
  pages     = "65--72",
  year      = {2024},
  doi       = {10.18653/v1/2024.scalellm-1.5}
}

@INPROCEEDINGS{Sevastjanova2023-lq,
  title     = "Visual comparison of text sequences generated by large language
               models",
  author    = "Sevastjanova, Rita and Vogelbacher, Simon and Spitz, Andreas and
               Keim, Daniel and El-Assady, Mennatallah",
  booktitle = "2023 IEEE Visualization in Data Science (VDS)",
  publisher = "IEEE",
  pages     = "11--20",
  month     =  oct,
  year      =  2023,
  language  = "en",
  doi       = {10.1109/vds60365.2023.00007}
}

@ARTICLE{Munz2022-bg,
  title     = "Visualization-based improvement of neural machine translation",
  author    = "Munz, Tanja and Väth, Dirk and Kuznecov, Paul and Vu, Ngoc Thang
               and Weiskopf, Daniel",
  journal   = "Computers \& Graphics",
  publisher = {Elsevier},
  volume    =  103,
  pages     = "45--60",
  month     =  apr,
  year      =  2022,
  doi       = {10.1016/j.cag.2021.12.003}
}

@INPROCEEDINGS{Strobelt2021-ag,
  title     = "{LMdiff}: A visual diff tool to compare language models",
  author    = {Strobelt, Hendrik and Hoover, Benjamin and Satyanarayan, Arvind
               and Gehrmann, Sebastian},
  booktitle = "Proceedings of the 2021 Conference on Empirical Methods in
               Natural Language Processing: System Demonstrations",
  publisher = "Association for Computational Linguistics",
  pages     = {96--105},
  month     =  nov,
  year      =  2021,
  doi       = {10.18653/v1/2021.emnlp-demo.12}
}

@MISC{Brockett2015-tq,
  title    = "User-modifiable word lattice display for editing documents and
              search queries",
  author   = "Brockett, Christopher John and Dolan, William Brennan",
  journal  = "US Patent",
  number   =  8972240,
  month    =  mar,
  year     =  2015
}

@inproceedings{brathvisualizing,
  title     = {Visualizing Textual Distributions of Repeated {LLM} Responses to Characterize {LLM} Knowledge},
  author    = {Brath, Richard and Bradley, Adam and Jonker, David},
  booktitle = {Proceedings of the VIS Workshop on NLP meets Visualization (NLVIZ)},
  year      = {2023},
  url       = {https://uncharted.software/assets/visualizing-textual-distributions-of-repeated-LLM-responses.pdf}
}

@inproceedings{brath2023visualizing,
  title     = {Visualizing {LLM} Text Style Transfer: Visually Dissecting How to Talk Like a Pirate},
  author    = {Brath, Richard and Bradley, Adam and Jonker, David},
  booktitle = {Proceedings of the VIS Workshop on NLP meets Visualization (NLVIZ)},
  year      = {2023},
  url       = {https://uncharted.software/assets/visualizing-LLM-text-style-transfer.pdf}
}

@ARTICLE{van-Ham2009-me,
  title     = "Mapping text with phrase nets",
  author    = "van Ham, Frank and Wattenberg, Martin and Viégas, Fernanda B",
  journal   = "IEEE Trans. Vis. Comput. Graph.",
  publisher = {IEEE},
  volume    =  15,
  number    =  6,
  pages     = "1169--1176",
  doi       = {10.1109/TVCG.2009.165},
  month     =  nov,
  year      =  2009,
  language  = "en"
}

@inproceedings{kuhn2023semantic,
  title     = {Semantic Uncertainty: Linguistic Invariances for Uncertainty Estimation in Natural Language Generation},
  author    = {Kuhn, Lorenz and Gal, Yarin and Farquhar, Sebastian},
  booktitle = {International Conference on Learning Representations},
  year      = {2023},
  url       = {https://openreview.net/forum?id=VD-AYtP0dve},
  eprint    = {2302.09664},
  archivePrefix = {arXiv},
  primaryClass  = {cs.CL},
  doi       = {10.48550/arXiv.2302.09664}
}

@inproceedings{collins2006leveraging,
  title={Leveraging uncertainty visualization to enhance multilingual chat},
  author={Collins, Christopher and Penn, Gerald},
  booktitle={Proceedings of the CSCW},
  year={2006}
}

@inproceedings{10.1145/3563657.3596138,
author = {Zamfirescu-Pereira, J.D. and Wei, Heather and Xiao, Amy and Gu, Kitty and Jung, Grace and Lee, Matthew G and Hartmann, Bjoern and Yang, Qian},
title = {Herding AI Cats: Lessons from Designing a Chatbot by Prompting GPT-3},
year = {2023},
isbn = {9781450398930},
publisher = {Association for Computing Machinery},
address = {New York, NY, USA},
url = {https://doi.org/10.1145/3563657.3596138},
doi = {10.1145/3563657.3596138},
booktitle = {Proceedings of the 2023 ACM Designing Interactive Systems Conference},
pages = {2206–2220},
numpages = {15},
location = {Pittsburgh, PA, USA},
series = {DIS '23}
}

@inproceedings{10.1145/3544548.3581388,
author = {Zamfirescu-Pereira, J.D. and Wong, Richmond Y. and Hartmann, Bjoern and Yang, Qian},
title = {Why Johnny Can’t Prompt: How Non-AI Experts Try (and Fail) to Design LLM Prompts},
year = {2023},
isbn = {9781450394215},
publisher = {Association for Computing Machinery},
address = {New York, NY, USA},
url = {https://doi.org/10.1145/3544548.3581388},
doi = {10.1145/3544548.3581388},
booktitle = {Proceedings of the 2023 CHI Conference on Human Factors in Computing Systems},
articleno = {437},
numpages = {21},
location = {Hamburg, Germany},
series = {CHI '23}
}

@misc{MaxRead,
  author = {Read, Max},
  title = {Who is Elara Voss?},
  howpublished = {\url{https://maxread.substack.com/p/who-is-elara-voss}},
  year = {2024},
  note = {Accessed: 2025-11-19}
}

@inproceedings{hopkins2023can,
  title={Can llms generate random numbers? evaluating llm sampling in controlled domains},
  author={Hopkins, Aspen K and Renda, Alex and Carbin, Michael},
  booktitle={ICML 2023 workshop: sampling and optimization in discrete space},
  year={2023}
}

@misc{zhang2025noveltybench,
  title         = {{NoveltyBench}: Evaluating Language Models for Humanlike Diversity},
  author        = {Zhang, Yiming and Diddee, Harshita and Holm, Susan and Liu,
                   Hanchen and Liu, Xinyue and Samuel, Vinay and Wang, Barry and
                   Ippolito, Daphne},
  year          = {2025},
  eprint        = {2504.05228},
  archivePrefix = {arXiv},
  primaryClass  = {cs.CL},
  doi           = {10.48550/arXiv.2504.05228},
  url           = {https://arxiv.org/abs/2504.05228}
}

@article{reinhart2025LMs,
  title={Do {LLMs} write like humans? Variation in grammatical and rhetorical styles},
  author={Reinhart, Alex and Markey, Ben and Laudenbach, Michael and Pantusen, Kachatad and Yurko, Ronald and Weinberg, Gordon and Brown, David West},
  journal={Proceedings of the National Academy of Sciences},
  volume={122},
  number={8},
  pages={e2422455122},
  year={2025},
  publisher={National Academy of Sciences},
  doi={10.1073/pnas.2422455122}
}

@inproceedings{10.1145/3706598.3713564,
author = {Agarwal, Dhruv and Naaman, Mor and Vashistha, Aditya},
title = {AI Suggestions Homogenize Writing Toward Western Styles and Diminish Cultural Nuances},
year = {2025},
isbn = {9798400713941},
publisher = {Association for Computing Machinery},
address = {New York, NY, USA},
url = {https://doi.org/10.1145/3706598.3713564},
doi = {10.1145/3706598.3713564},
booktitle = {Proceedings of the 2025 CHI Conference on Human Factors in Computing Systems},
articleno = {1117},
numpages = {21},
location = {
},
series = {CHI '25}
}

@article{braunUsingThematicAnalysis2006,
  title = {Using Thematic Analysis in Psychology},
  author = {Braun, Virginia and Clarke, Victoria},
  year = {2006},
  month = jan,
  journal = {Qualitative Research in Psychology},
  volume = {3},
  number = {2},
  pages = {77--101},
  publisher = {Routledge},
  issn = {1478-0887},
  doi = {10.1191/1478088706qp063oa},
  urldate = {2025-08-31}
}

@article{morseCriticalAnalysisStrategies2015,
  title = {Critical {{Analysis}} of {{Strategies}} for {{Determining Rigor}} in {{Qualitative Inquiry}}},
  author = {Morse, Janice M.},
  year = {2015},
  month = sep,
  journal = {Qualitative Health Research},
  volume = {25},
  number = {9},
  pages = {1212--1222},
  issn = {1049-7323},
  doi = {10.1177/1049732315588501},
  langid = {english},
  pmid = {26184336}
}

@INPROCEEDINGS{6758829,
  author={Heimerl, Florian and Lohmann, Steffen and Lange, Simon and Ertl, Thomas},
  booktitle={2014 47th Hawaii International Conference on System Sciences},
  title={Word Cloud Explorer: Text Analytics Based on Word Clouds},
  year={2014},
  pages={1833-1842},
  doi={10.1109/HICSS.2014.231}}

@article{wattenberg2008word,
  title={The word tree, an interactive visual concordance},
  author={Wattenberg, Martin and Vi{\'e}gas, Fernanda B},
  journal={IEEE Transactions on Visualization and Computer Graphics},
  volume={14},
  number={6},
  pages={1221--1228},
  year={2008},
  publisher={IEEE},
  doi={10.1109/TVCG.2008.172}
}

@inproceedings{Furnas1986TheFV,
  title        = {Generalized fisheye views},
  author       = {Furnas, George W.},
  booktitle    = {Proceedings of the SIGCHI Conference on Human Factors in Computing Systems},
  year         = {1986},
  pages        = {16--23},
  publisher    = {ACM},
  doi          = {10.1145/22627.22342}
}

@misc{Swoopes2025stochasticity,
  title         = {The Impact of Revealing Large Language Model Stochasticity on Trust, Reliability, and Anthropomorphization},
  author        = {Swoopes, Chelse and Holloway, Tyler and Glassman, Elena L.},
  year          = {2025},
  eprint        = {2503.16114},
  archivePrefix = {arXiv},
  primaryClass  = {cs.HC},
  doi           = {10.48550/arXiv.2503.16114},
  url           = {https://arxiv.org/abs/2503.16114}
}

@article{Bach2016-timecurves,
  author = {Bach, Benjamin and Shi, Conglei and Heulot, Nicolas and Madhyastha, Tara and Grabowski, Tom and Dragicevic, Pierre},
  title = {Time Curves: Folding Time to Visualize Patterns of Temporal Evolution in Data},
  year = {2016},
  publisher = {IEEE Educational Activities Department},
  address = {USA},
  volume = {22},
  number = {1},
  journal = {IEEE Transactions on Visualization and Computer Graphics},
  month = jan,
  pages = {559--568},
  doi = {10.1109/TVCG.2015.2467851},
  url = {https://doi.org/10.1109/TVCG.2015.2467851}
}

@misc{jiang2025artificialhivemindopenendedhomogeneity,
  title={Artificial Hivemind: The Open-Ended Homogeneity of Language Models (and Beyond)},
  author={Liwei Jiang and Yuanjun Chai and Margaret Li and Mickel Liu and Raymond Fok and Nouha Dziri and Yulia Tsvetkov and Maarten Sap and Alon Albalak and Yejin Choi},
  year={2025},
  eprint={2510.22954},
  archivePrefix={arXiv},
  primaryClass={cs.CL},
  doi={10.48550/arXiv.2510.22954},
  url={https://arxiv.org/abs/2510.22954}
}

@article{wilcoxon1945,
  title={Individual comparisons by ranking methods},
  author={Wilcoxon, Frank},
  journal={Biometrics Bulletin},
  volume={1},
  number={6},
  pages={80--83},
  year={1945},
  doi={10.2307/3001968}
}

@misc{wattenberg_elara_2025,
  author       = {Laura Wattenberg},
  title        = {2025 Name of the Year Is Elara, the Favorite Name of AI},
  howpublished = {\url{https://namerology.com/2025/12/15/2025-name-of-the-year-is-elara-the-favorite-name-of-ai/}},
  year         = {2025},
  note         = {Accessed: 2026-03-19}
}

@inproceedings{10.1145/3626772.3657914,
author = {Trippas, Johanne R. and Al Lawati, Sara Fahad Dawood and Mackenzie, Joel and Gallagher, Luke},
title = {What do Users Really Ask Large Language Models? An Initial Log Analysis of Google Bard Interactions in the Wild},
year = {2024},
isbn = {9798400704314},
publisher = {Association for Computing Machinery},
address = {New York, NY, USA},
url = {https://doi.org/10.1145/3626772.3657914},
doi = {10.1145/3626772.3657914},
booktitle = {Proceedings of the 47th International ACM SIGIR Conference on Research and Development in Information Retrieval},
pages = {2703–2707},
numpages = {5},
location = {Washington DC, USA},
series = {SIGIR '24}
}

@article{Ma2023UnderstandingTB,
  title   = {Understanding the Benefits and Challenges of Using Large Language Model-based Conversational Agents for Mental Well-being Support},
  author  = {Ma, Zilin and Mei, Yiyang and Su, Zhaoyuan},
  journal = {AMIA Annual Symposium Proceedings},
  year    = {2023},
  volume  = {2023},
  pages   = {1105--1114},
  pmid    = {38222348},
  pmcid   = {PMC10785945},
  url     = {https://www.ncbi.nlm.nih.gov/pmc/articles/PMC10785945/}
}

@article{zhang2025exploring,
  title={Exploring the role of large language models in the scientific method: from hypothesis to discovery},
  author={Zhang, Yanbo and Khan, Sumeer A and Mahmud, Adnan and Yang, Huck and Lavin, Alexander and Levin, Michael and Frey, Jeremy and Dunnmon, Jared and Evans, James and Bundy, Alan and others},
  journal={npj Artificial Intelligence},
  volume={1},
  number={1},
  pages={14},
  year={2025},
  publisher={Nature Publishing Group},
  doi={10.1038/s44387-025-00019-5}
}

@book{young1989token,
  title={Token passing: a simple conceptual model for connected speech recognition systems},
  author={Young, Stephen John and Russell, NH and Thornton, JH Simon},
  year={1989},
  publisher={Cambridge University Engineering Department Cambridge, UK}
}

@inproceedings{wu2022ai,
  title={{AI} Chains: Transparent and Controllable Human-{AI} Interaction by Chaining Large Language Model Prompts},
  author={Wu, Tongshuang and Terry, Michael and Cai, Carrie Jun},
  booktitle={Proceedings of the 2022 CHI Conference on Human Factors in Computing Systems},
  pages={1--22},
  year={2022},
  publisher={ACM},
  doi={10.1145/3491102.3517582}
}

@article{acharya2025agentic,
  title={Agentic {AI}: Autonomous Intelligence for Complex Goals: A Comprehensive Survey},
  author={Acharya, Deepak Bhaskar and Kuppan, Karthigeyan and Divya, B},
  journal={IEEE Access},
  volume={13},
  pages={18912--18936},
  year={2025},
  publisher={IEEE},
  doi={10.1109/ACCESS.2025.3532853}
}

@misc{pang2025interactive,
  title={Interactive Reasoning: Visualizing and Controlling Chain-of-Thought Reasoning in Large Language Models},
  author={Pang, Rock Yuren and Feng, K. J. Kevin and Feng, Shangbin and Li, Chu and Shi, Weijia and Tsvetkov, Yulia and Heer, Jeffrey and Reinecke, Katharina},
  year={2025},
  eprint={2506.23678},
  archivePrefix={arXiv},
  primaryClass={cs.HC},
  doi={10.48550/arXiv.2506.23678},
  url={https://arxiv.org/abs/2506.23678}
}

@inproceedings{shneiderman1996eyes,
  title={The Eyes Have It: A Task by Data Type Taxonomy for Information Visualizations},
  author={Shneiderman, Ben},
  booktitle={Proceedings 1996 IEEE Symposium on Visual Languages},
  pages={336--343},
  year={1996},
  publisher={IEEE},
  doi={10.1109/VL.1996.545307}
}

@inproceedings{pirolli2005sensemaking,
  title={The sensemaking process and leverage points for analyst technology as identified through cognitive task analysis},
  author={Pirolli, Peter and Card, Stuart},
  booktitle={Proceedings of international conference on intelligence analysis},
  volume={5},
  number={1},
  pages={2--4},
  year={2005},
  organization={McLean, VA, USA}
}

@misc{ wiki:Signs_of_AI_writing,
  author = "{Wikipedia contributors}",
  title = "Wikipedia:Signs of AI writing --- {W}ikipedia{,} The Free Encyclopedia",
  year = "2026",
  url = "https://en.wikipedia.org/w/index.php?title=Wikipedia:Signs_of_AI_writing",
  note = "[Online; accessed 30-March-2026]"
}

@article{10.1145/3652028,
author = {Spinner, Thilo and Kehlbeck, Rebecca and Sevastjanova, Rita and St\"{a}hle, Tobias and Keim, Daniel A. and Deussen, Oliver and El-Assady, Mennatallah},
title = {-generAItor: Tree-in-the-loop Text Generation for Language Model Explainability and Adaptation},
year = {2024},
issue_date = {June 2024},
publisher = {Association for Computing Machinery},
address = {New York, NY, USA},
volume = {14},
number = {2},
issn = {2160-6455},
url = {https://doi.org/10.1145/3652028},
doi = {10.1145/3652028},
journal = {ACM Trans. Interact. Intell. Syst.},
month = jun,
articleno = {14},
numpages = {32}
}

\clearpage
\appendix
\crefalias{section}{appendix}
\setcounter{figure}{0}
\renewcommand{\thefigure}{A.\arabic{figure}}
\setcounter{algorithm}{0}
\renewcommand{\thealgorithm}{A.\arabic{algorithm}}

\section{Token similarity algorithm}
\label{sec:alg-similarity}

\begin{algorithm}
  \caption{Token similarity score used during the merge step. For stopwords, similarity is based on neighboring token embeddings rather than the tokens themselves. A positional penalty discourages merging tokens that are far apart in the original sentence.}
\begin{algorithmic}
\Function{CosSim}{wordA, wordB}
  \State embA $\gets$ Embed(wordA)
  \State embB $\gets$ Embed(wordB)
  \State \Return CosSim(embA, embB)
\EndFunction

\Function{Similarity}{wordA, wordB}

  \If{\Call{isStopWord}{wordA} \Call{isStopWord}{wordB}}
      \State simPrev $\gets$ \Call{CosSim}{wordBeforeA, wordBeforeB}
      \State simNext $\gets$ \Call{CosSim}{wordAfterA, wordAfterB}
      \State score $\gets$ (simPrev + simNext) / 2
  \Else
      \State score $\gets$ CosSim(wordA, wordB)
  \EndIf

  \State positionalPenalty $\gets$ abs(wordA.idx - wordB.idx)/20
  \State score $\gets$ score - positionalPenalty

  \State \Return score
\EndFunction
\end{algorithmic}
\label{alg:similarity}
\end{algorithm}
\setcounter{figure}{1}

\begin{figure}[h]
  \centering
  \begin{subfigure}[t]{\linewidth}
    \includegraphics[width=\linewidth]{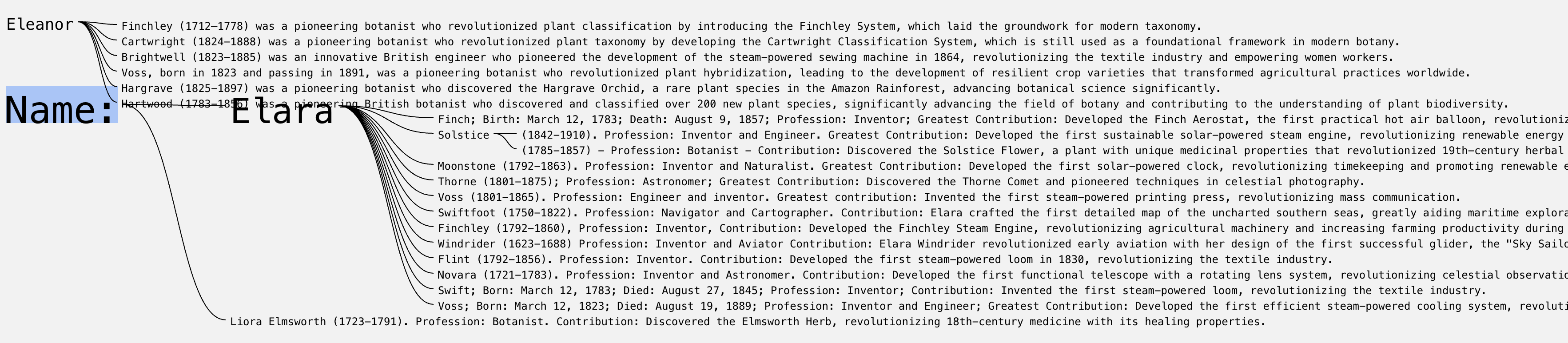}
    \caption{Word tree prototype.}
    \label{fig:wordtree}
  \end{subfigure}
  \vspace{\smallskipamount}
  \begin{subfigure}[t]{\linewidth}
    \includegraphics[width=\linewidth]{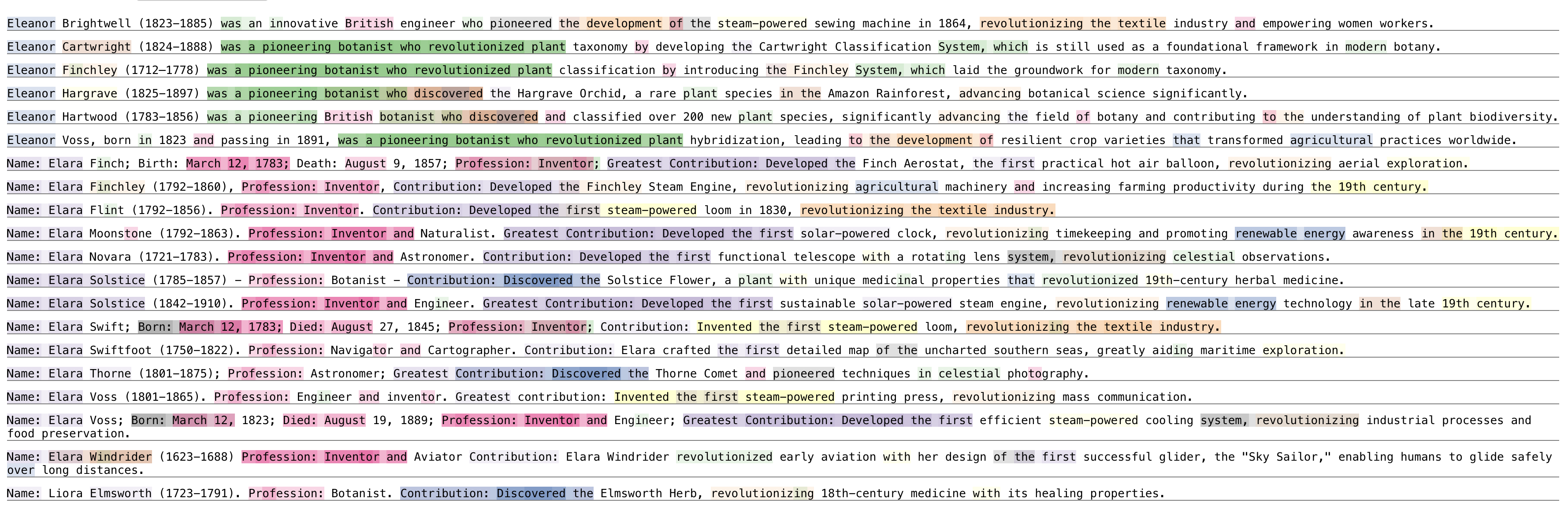}
    \caption{Color-coded span highlighting prototype.}
    \label{fig:highlights}
  \end{subfigure}
  \vspace{\smallskipamount}
  \begin{subfigure}[t]{\linewidth}
    \includegraphics[width=\linewidth]{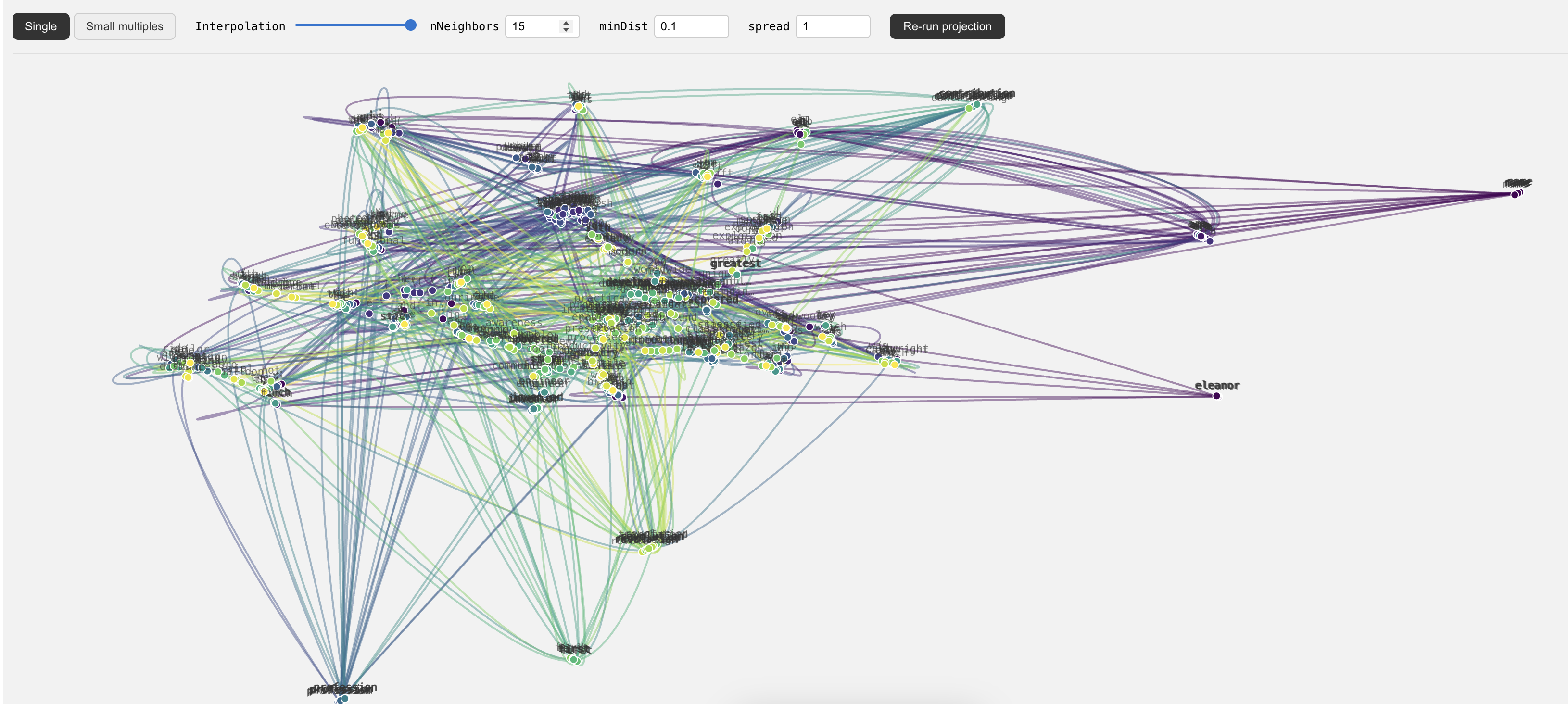}
    \caption{Time Curves-style prototype~\cite{Bach2016-timecurves}.}
    \label{fig:timegraph}
  \end{subfigure}
  \caption{During our iterative design process, we prototyped several alternative visual encodings before settling on the graph-based approach: (a) word tree, (b) color-coded span highlighting, (c) Time Curves-style representation. Early lightweight tests indicated that these encodings did not adequately surface branching and merging structure across outputs.}
  \label{fig:alt-encodings}
\end{figure}

\begin{figure*}[h]
  \centering
  \includegraphics[width=\linewidth]{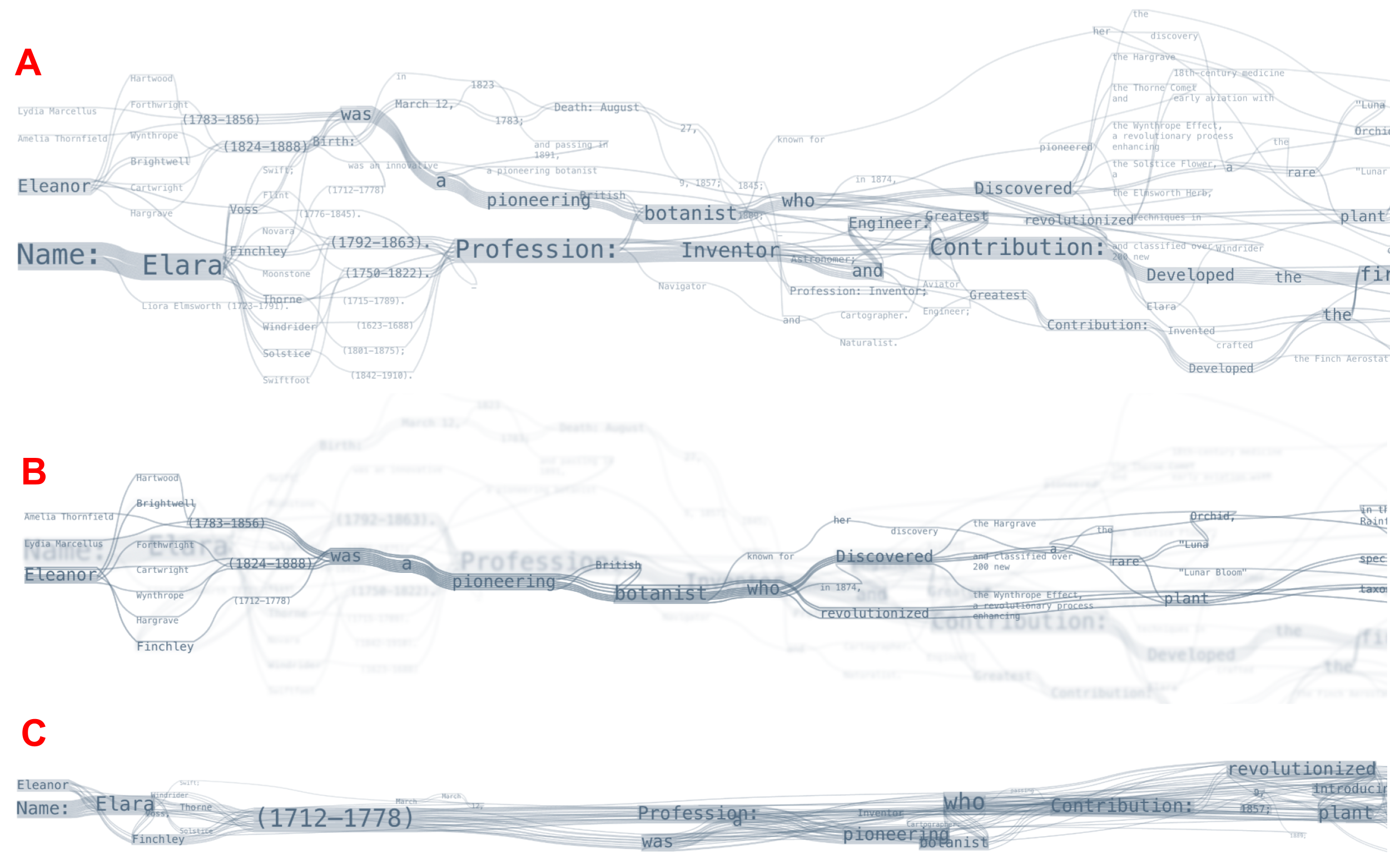}
  \caption{A comparison of graph filtering methods. A: original generations (no filtering). B: Filtering by selecting ``botanist'' to see the subset of the graph containing that word. C: Lowering the semantic similarity merge threshold to .2. }
  \label{fig:compare_filter}
\end{figure*}


\begin{figure}[h]
  \centering
  \includegraphics[width=\linewidth]{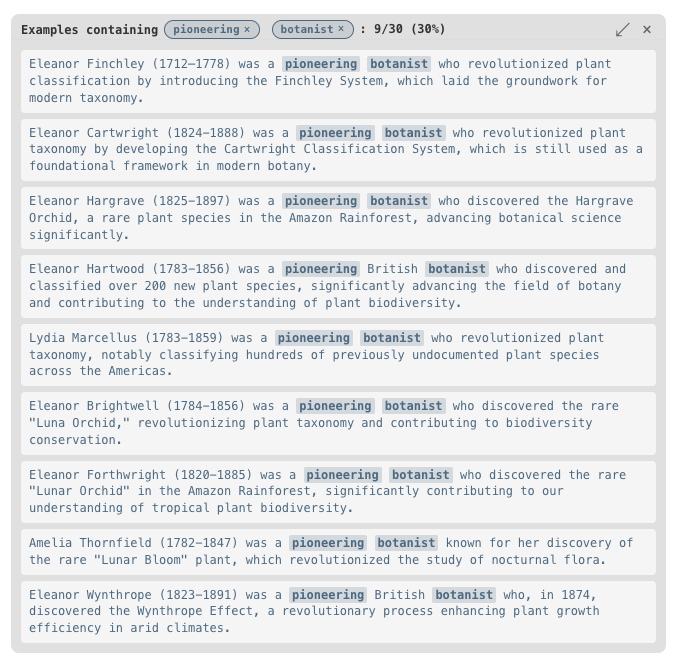}
  \caption{Users can select one or more nodes in the graph to filter the view to generations containing those words or phrases.}
  \label{fig:view_subset}
\end{figure}

\section{Formative study protocol}
\label{sec:protocol}
Each session followed a semi-structured protocol and lasted 30 to 45 minutes. Sessions were recorded with consent and supplemented with researcher notes. The protocol was tailored to each participant's background and use cases but followed a consistent structure:

\paragraph{1. Introduction (2 to 3 min).} The facilitator explained the study's purpose: to understand how LM researchers currently assess, address, and leverage the stochasticity of models, and to identify potential use cases for a graph-based visualization of LM output distributions.

\paragraph{2. Background and semi-structured discussion (10 to 15 min).} Participants described their current research focus and how they typically use LMs. Questions were adapted to each participant's context but drew from the following:
\begin{itemize}
    \item \textit{``What kinds of NLP/LLM/AI problems are you working on?''}
    \item \textit{``Do your tasks have constrained outputs (e.g., multiple choice), or are they open-ended?''}
    \item \textit{``How do you evaluate the LM on a novel, open-ended task?''}
    \item \textit{``How do you currently interact with LMs when testing early research hypotheses, or using LMs for a task?''} (e.g., chatbot interface, notebook, eval pipelines)
    \item \textit{``When are multiple outputs important?''}
    \item \textit{``When are multiple \textbf{sets} of outputs important?''}
    \item \textit{``Do you call models multiple times to check for consistency or for other reasons?''}
    \item \textit{``Do you have situations where you examine many outputs at once?''}
\end{itemize}
When possible, the facilitator asked participants to ground their answers in a specific, recent example rather than generalizations.

\paragraph{3. Prototype demo (5 min).} Participants viewed an early prototype of the visualization tool. The facilitator walked through a preloaded example, demonstrating the graph layout, node selection and filtering, and comparison mode.

\paragraph{4. Feedback (5 to 10 min).} Participants gave reactions to the prototype's clarity, utility, and perceived relevance (or lack thereof) to their own workflows. Sample prompts included:
\begin{itemize}
    \item \textit{``Is this clear? What is confusing?''}
    \item \textit{``Would this be useful for any of the tasks you described?''}
    \item \textit{``What is missing or would you change?''}
\end{itemize}

\paragraph{5. Use case brainstorming (5 to 10 min).} Participants were asked to imagine how the tool could fit into their current work, and discussed potential extensions, new features, or alternative analysis tasks the tool could support.

\paragraph{6. Wrap-up (2 min).} The facilitator asked for any remaining thoughts and thanked the participant.

\paragraph{Analysis.} Interview notes and recordings were coded to identify common workflow patterns for LM output inspection, recurrent themes in potential use cases, and specific design feedback. We used open coding followed by thematic synthesis, as described in \S\ref{sec:formative}.

\begin{figure}[h]
  \centering
  \begin{subfigure}[t]{\linewidth}
    \centering
    \includegraphics[width=\linewidth]{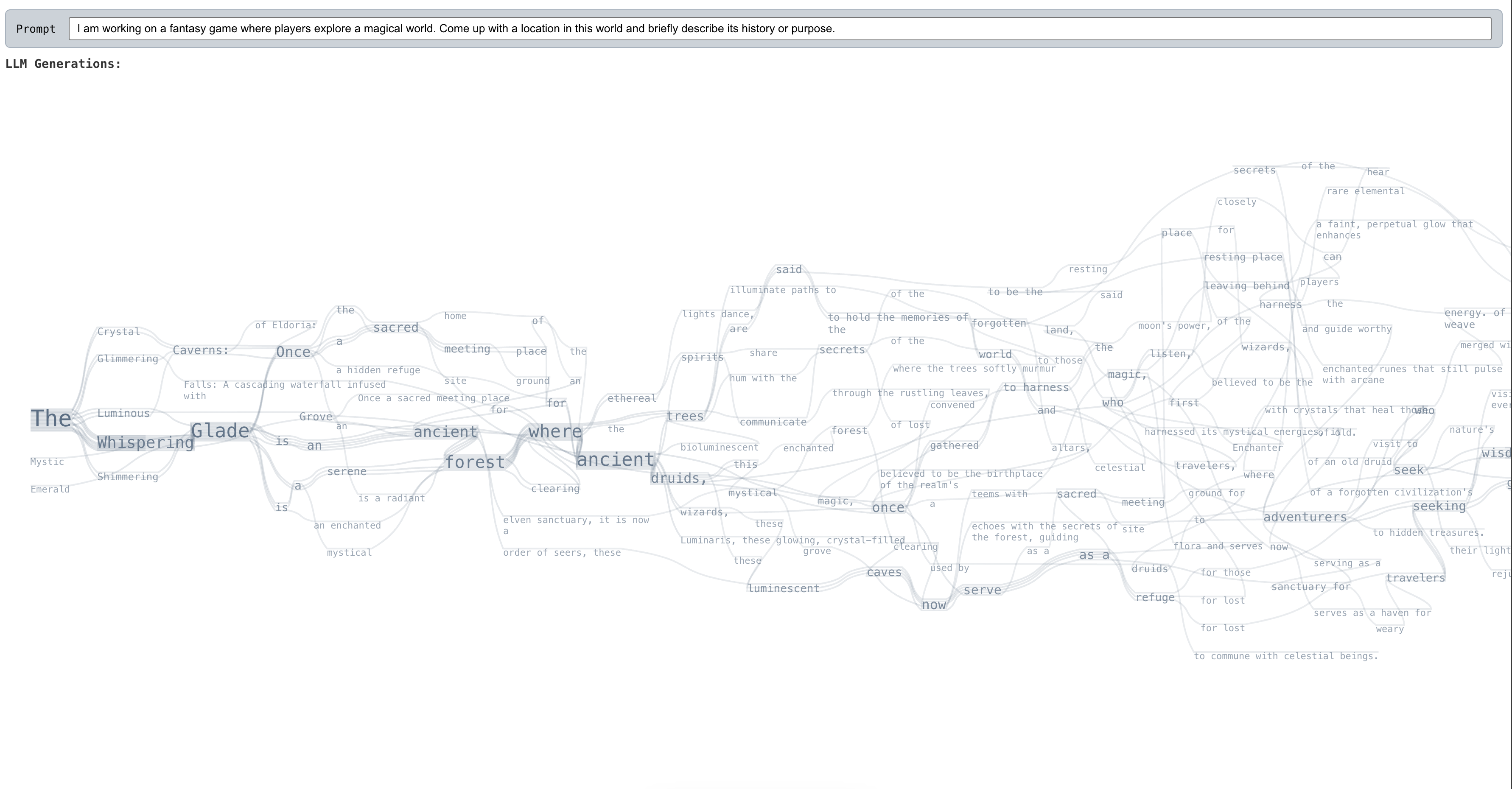}
    \subcaption{Graph interface.}
  \end{subfigure}
  \vspace{0.5em}
  \begin{subfigure}[t]{\linewidth}
    \centering
    \includegraphics[width=\linewidth]{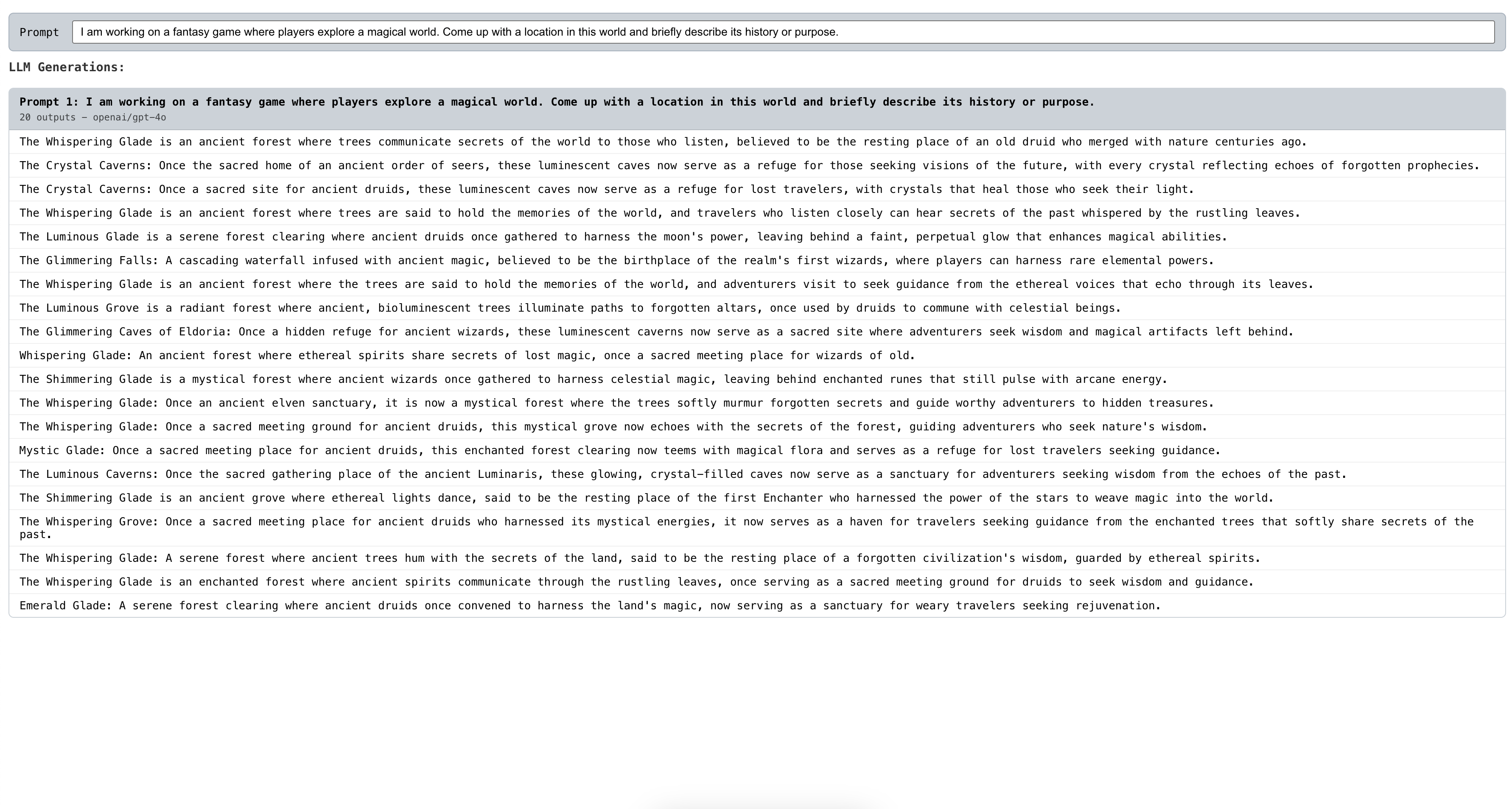}
    \subcaption{List interface.}
  \end{subfigure}
  \caption{Graph and list UIs for the single-distribution comprehension study. Participants viewed $k{=}20$ outputs from one prompt and answered a structured questionnaire about the distribution. The layout was the same in both cases.}
  \label{fig:user-study-single}
\end{figure}


\begin{figure}[h]
  \centering
  \begin{subfigure}[t]{\linewidth}
    \centering
    \includegraphics[width=\linewidth]{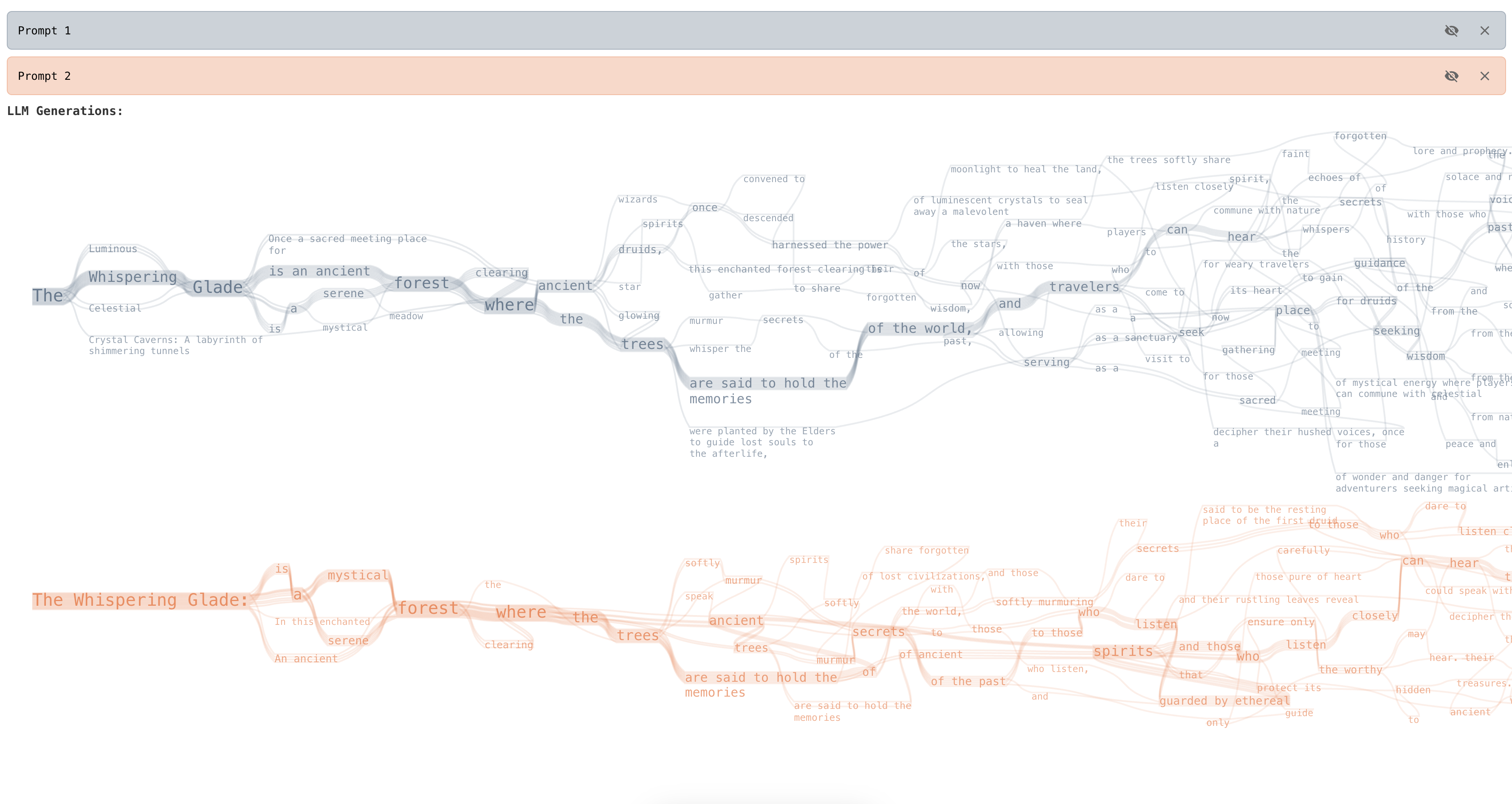}
    \subcaption{Graph interface.}
  \end{subfigure}
  \vspace{0.5em}
  \begin{subfigure}[t]{\linewidth}
    \centering
    \includegraphics[width=\linewidth]{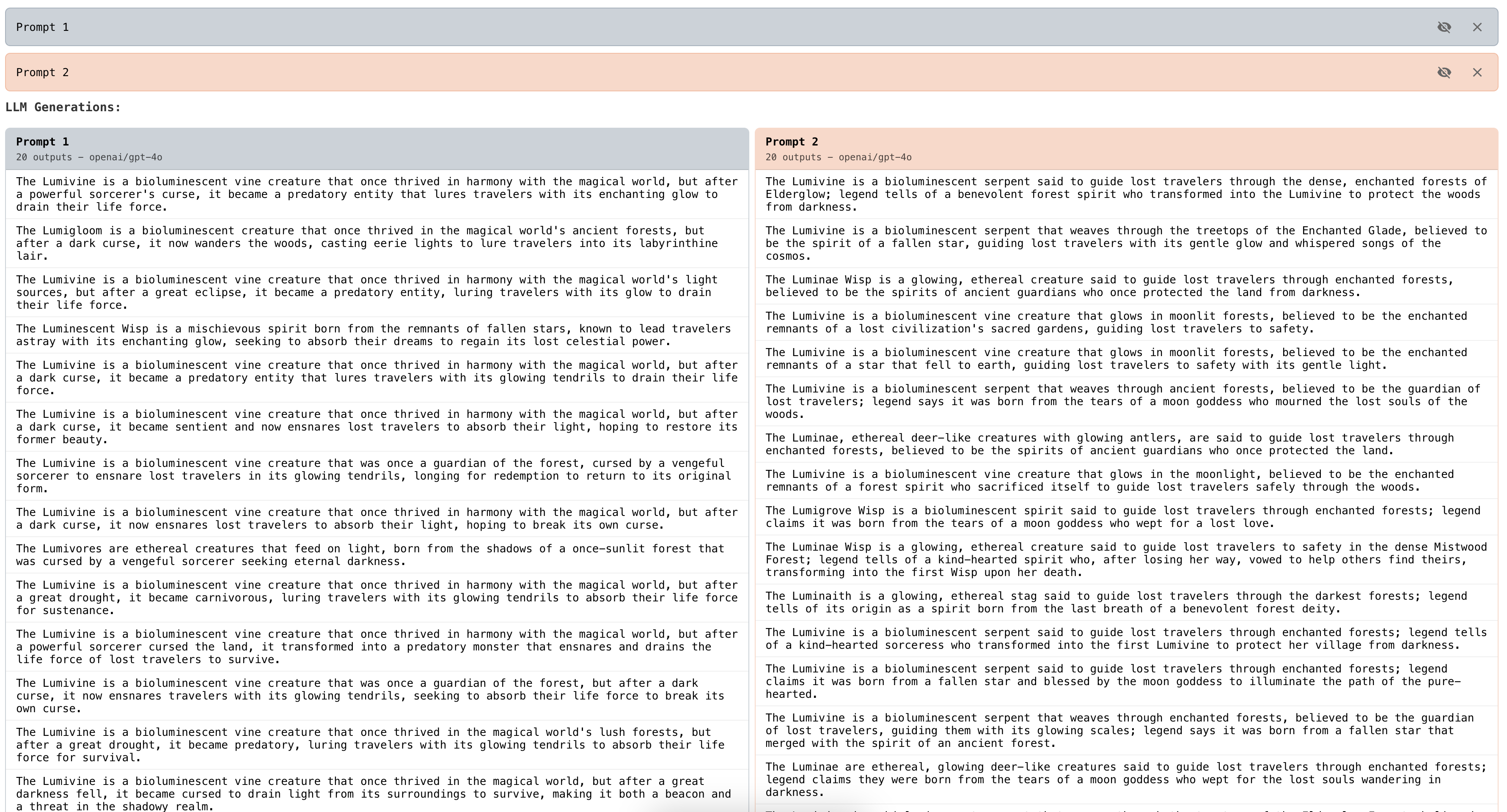}
    \subcaption{List interface.}
  \end{subfigure}
  \caption{Graph and list UIs for the diversity comparison and two-distribution comparison studies. The layout was the same in both cases; underlying data and task differed as described above.}
  \label{fig:user-study-diversity}
\end{figure}

\begin{figure}[h]
  \centering
  \includegraphics[width=\linewidth]{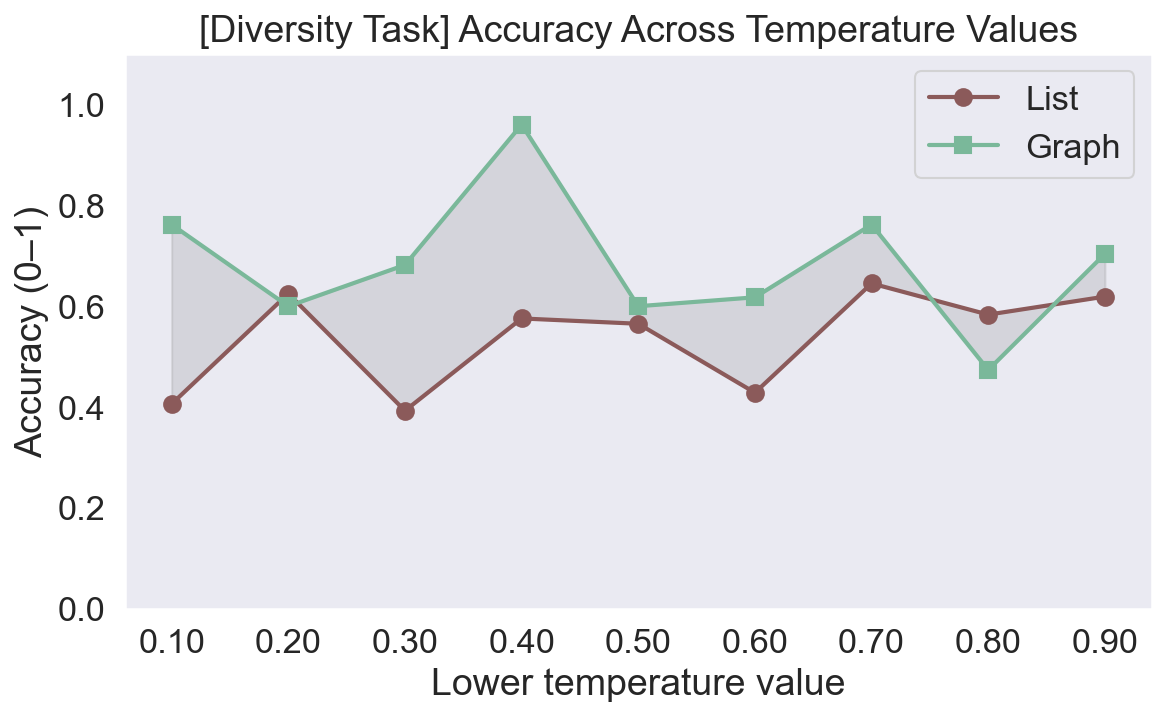}
  \caption{Accuracy for the diversity task by temperature bucket (i.e., diversity level of the outputs). The largest interface differences appear at low to medium temperature; interpretation is limited because diversity depends on the prompt as well as temperature.}
  \label{fig:diversity_accuracy_by_temp}
\end{figure}

\begin{figure}[h]
  \centering
  \includegraphics[width=\linewidth]{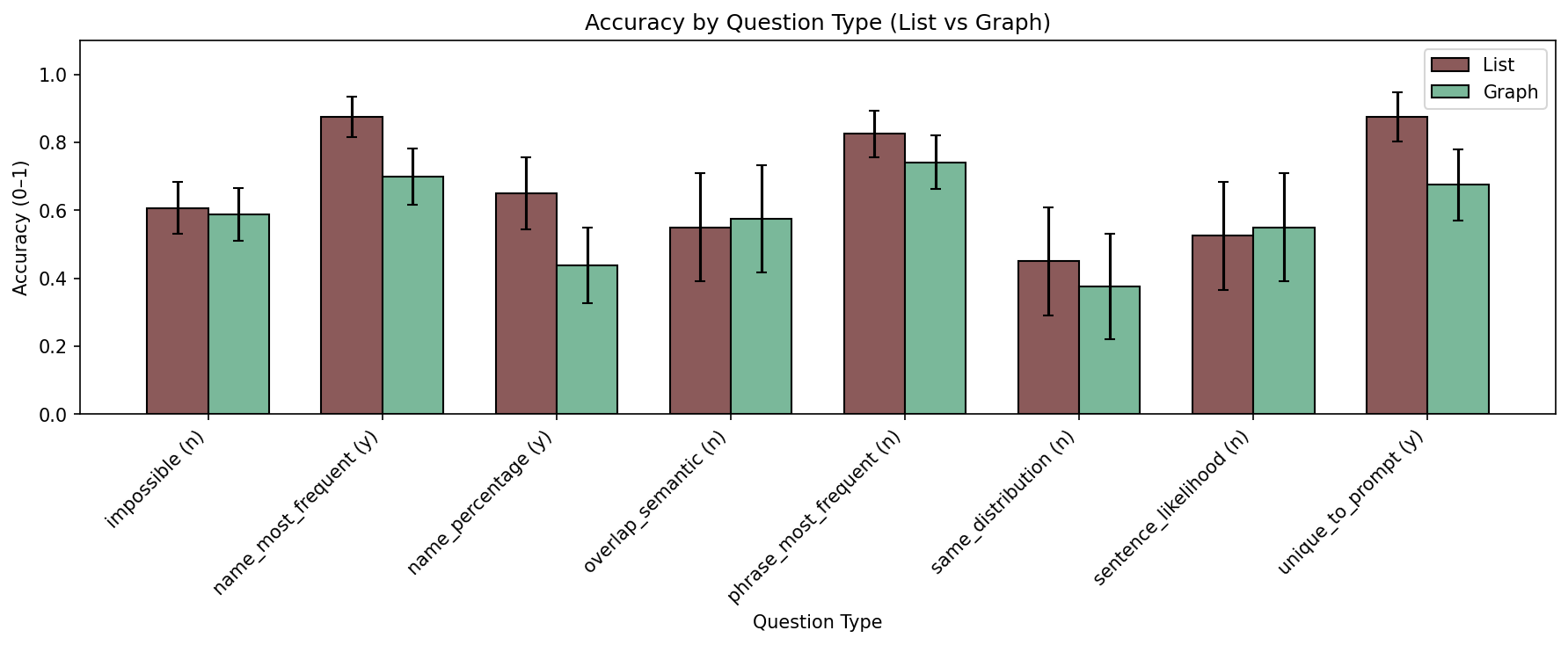}
  \caption{Breakdown of accuracy by question type for the two-distribution comparison task. List outperforms graph on detail-related questions (e.g., name frequency, phrase matching, percentage estimates).}
  \label{fig:compare-accuracy-by-qtype}
\end{figure}

\section{Post-Task Survey Protocol}
\label{sec:post-task-survey}
After completing tasks with each interface, participants completed a post-task survey. For each of the graph and list interfaces, participants answered the following questions on a 7-point Likert scale (strongly disagree to strongly agree):

\paragraph{Understanding model behavior.}
\begin{itemize}
  \item ``Using this interface, I understood how diverse (i.e., how narrow or broad) the output space was for a given prompt.''
  \item ``I understood what a typical (or `average') output looked like for a given prompt.''
  \item ``I had a good sense of what rare or unusual outputs looked like for a given prompt.''
  \item ``I felt I had seen enough of the model's behavior for a prompt to make a good decision.''
  \item ``I could see if there were recurring patterns in the outputs, and if so, what types.''
\end{itemize}

\paragraph{Workload and Effort (NASA TLX).}
\begin{itemize}
  \item ``Using this interface required a lot of mental effort.''
  \item ``I had to work hard to complete the task using this interface.''
  \item ``I felt frustrated while using this interface.''
  \item ``I felt rushed while using this interface.''
\end{itemize}

\paragraph{Usability and satisfaction.}
\begin{itemize}
  \item ``I found the interface easy to use.''
  \item ``I felt confident in the decisions I made using this interface.''
  \item ``I would want to use this interface for similar tasks in my own work.''
\end{itemize}

Participants also answered the following comparison questions on a 7-point scale anchored by the two interfaces (1 = graph, 7 = list):
\begin{itemize}
  \item Which interface made the task easier overall?
  \item Which interface felt more overwhelming to use?
  \item Which interface better supported understanding the range and distribution of outputs?
  \item Which interface made you feel more confident in your answers?
  \item Overall, which interface do you prefer?
\end{itemize}

\begin{figure}[t]
  \centering
  \small
  \renewcommand{\arraystretch}{1.0}
  \begin{tabular}{@{}p{3.8cm}>{\centering\arraybackslash}p{1cm}>{\centering\arraybackslash}p{1cm}>{\centering\arraybackslash}p{1cm}@{}}
  \toprule
  & Diversity & Single & Compare \\
  \midrule
  \multicolumn{4}{@{}l}{\textit{Understanding model behavior}} \\
  \midrule
  Understood diversity of output space & \cellcolor[rgb]{0.952,0.932,0.932} \shortstack{r=-0.11\\$p$=0.668} & \cellcolor[rgb]{0.843,0.916,0.881} \shortstack{r=0.30\\$p$=0.206} & \cellcolor[rgb]{0.868,0.930,0.900} \shortstack{r=0.25\\$p$=0.332} \\
  \midrule
  Understood typical output & \cellcolor[rgb]{0.997,0.998,0.998} \shortstack{r=0.01\\$p$=0.989} & \cellcolor[rgb]{0.933,0.904,0.904} \shortstack{r=-0.15\\$p$=0.502} & \cellcolor[rgb]{0.799,0.893,0.847} \shortstack{r=0.39\\$p$=0.142} \\
  \midrule
  Understood rare/unusual outputs & \cellcolor[rgb]{0.889,0.841,0.841} \shortstack{r=-0.25\\$p$=0.318} & \cellcolor[rgb]{0.923,0.959,0.941} \shortstack{r=0.15\\$p$=0.533} & \cellcolor[rgb]{0.794,0.890,0.843} \shortstack{r=0.40\\$p$=0.116} \\
  \midrule
  Seen enough for good decision & \cellcolor[rgb]{0.871,0.817,0.817} \shortstack{r=-0.28\\$p$=0.199} & \cellcolor[rgb]{0.828,0.755,0.755} \shortstack{r=-0.38\\$p$=0.090} & \cellcolor[rgb]{0.959,0.941,0.941} \shortstack{r=-0.09\\$p$=0.728} \\
  \midrule
  Could see recurring patterns & \cellcolor[rgb]{0.889,0.841,0.841} \shortstack{r=-0.25\\$p$=0.276} & \cellcolor[rgb]{0.865,0.808,0.808} \shortstack{r=-0.30\\$p$=0.218} & \cellcolor[rgb]{0.911,0.953,0.933} \shortstack{r=0.17\\$p$=0.483} \\
  \midrule
  \multicolumn{4}{@{}l}{\textit{Workload and Effort (NASA TLX)}} \\
  \midrule
  Required mental effort & \cellcolor[rgb]{0.996,0.998,0.997} \shortstack{r=0.01\\$p$=0.980} & \cellcolor[rgb]{0.903,0.863,0.863} \shortstack{r=-0.21\\$p$=0.356} & \cellcolor[rgb]{0.973,0.961,0.961} \shortstack{r=-0.06\\$p$=0.804} \\
  \midrule
  Had to work hard & \cellcolor[rgb]{0.982,0.990,0.986} \shortstack{r=0.03\\$p$=0.887} & \cellcolor[rgb]{0.997,0.998,0.997} \shortstack{r=0.01\\$p$=0.983} & \cellcolor[rgb]{0.898,0.855,0.855} \shortstack{r=-0.22\\$p$=0.336} \\
  \midrule
  Felt frustrated & \cellcolor[rgb]{0.974,0.962,0.962} \shortstack{r=-0.06\\$p$=0.819} & \cellcolor[rgb]{0.877,0.825,0.825} \shortstack{r=-0.27\\$p$=0.239} & \cellcolor[rgb]{0.967,0.982,0.975} \shortstack{r=0.06\\$p$=0.820} \\
  \midrule
  Felt rushed & \cellcolor[rgb]{0.983,0.975,0.975} \shortstack{r=-0.04\\$p$=0.895} & \cellcolor[rgb]{0.998,0.999,0.999} \shortstack{r=0.00\\$p$=1.000} & \cellcolor[rgb]{0.880,0.936,0.909} \shortstack{r=0.23\\$p$=0.481} \\
  \midrule
  \multicolumn{4}{@{}l}{\textit{Usability and satisfaction}} \\
  \midrule
  Interface easy to use & \cellcolor[rgb]{0.981,0.972,0.972} \shortstack{r=-0.04\\$p$=0.856} & \cellcolor[rgb]{0.943,0.919,0.919} \shortstack{r=-0.13\\$p$=0.563} & \cellcolor[rgb]{0.933,0.964,0.949} \shortstack{r=0.13\\$p$=0.563} \\
  \midrule
  Felt confident in decisions & \cellcolor[rgb]{0.892,0.847,0.847} \shortstack{r=-0.24\\$p$=0.294} & \cellcolor[rgb]{0.915,0.879,0.879} \shortstack{r=-0.19\\$p$=0.385} & \cellcolor[rgb]{0.854,0.792,0.792} \shortstack{r=-0.32\\$p$=0.213} \\
  \midrule
  Would use for own work & \cellcolor[rgb]{0.901,0.859,0.859} \shortstack{r=-0.22\\$p$=0.378} & \cellcolor[rgb]{0.953,0.975,0.964} \shortstack{r=0.09\\$p$=0.660} & \cellcolor[rgb]{0.846,0.781,0.781} \shortstack{r=-0.34\\$p$=0.204} \\
  \bottomrule
  \end{tabular}
  \caption{Post-task Likert scale: Wilcoxon signed-rank test (graph vs.\ list) for each question across all three studies. Effect size $r$ (rank-biserial correlation) and $p$-value. No differences reached statistical significance ($p > 0.05$). Positive $r$ favors graph; negative $r$ favors list. Colors: purple (list favored) to green (graph favored).}
  \label{fig:likert_post_survey}
\end{figure}

\begin{figure}[h]
  \centering
  \begin{subfigure}[t]{\linewidth}
    \centering
    \includegraphics[width=\linewidth]{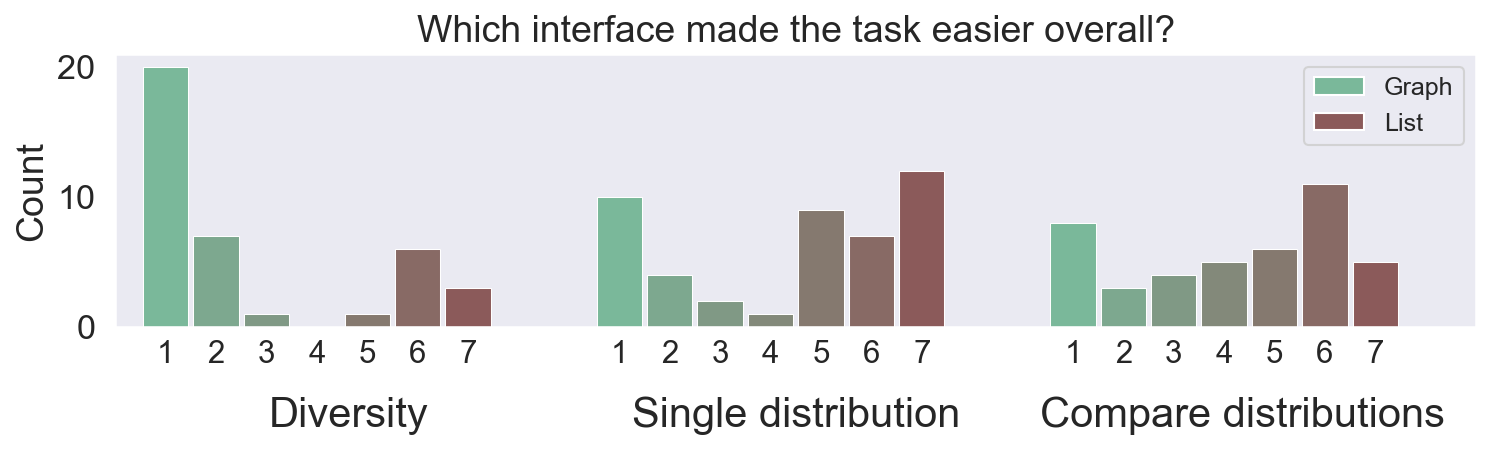}
  \end{subfigure}
  \vspace{0.5em}
  \begin{subfigure}[t]{\linewidth}
    \centering
    \includegraphics[width=\linewidth]{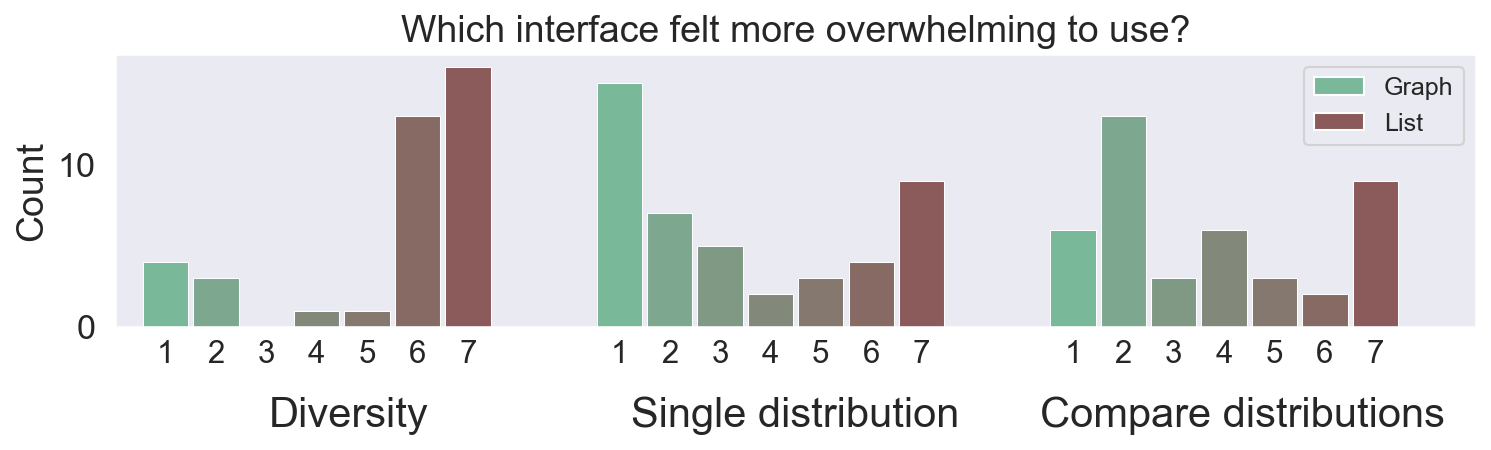}
  \end{subfigure}
  \vspace{0.5em}
  \begin{subfigure}[t]{\linewidth}
    \centering
    \includegraphics[width=\linewidth]{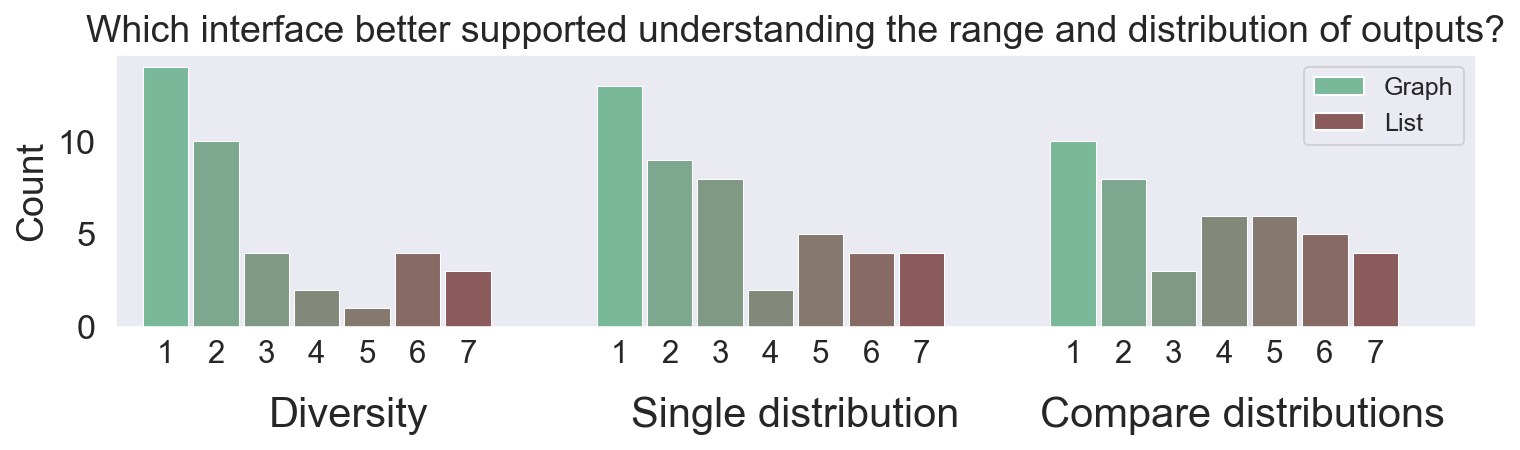}
  \end{subfigure}
  \vspace{0.5em}
  \begin{subfigure}[t]{\linewidth}
    \centering
    \includegraphics[width=\linewidth]{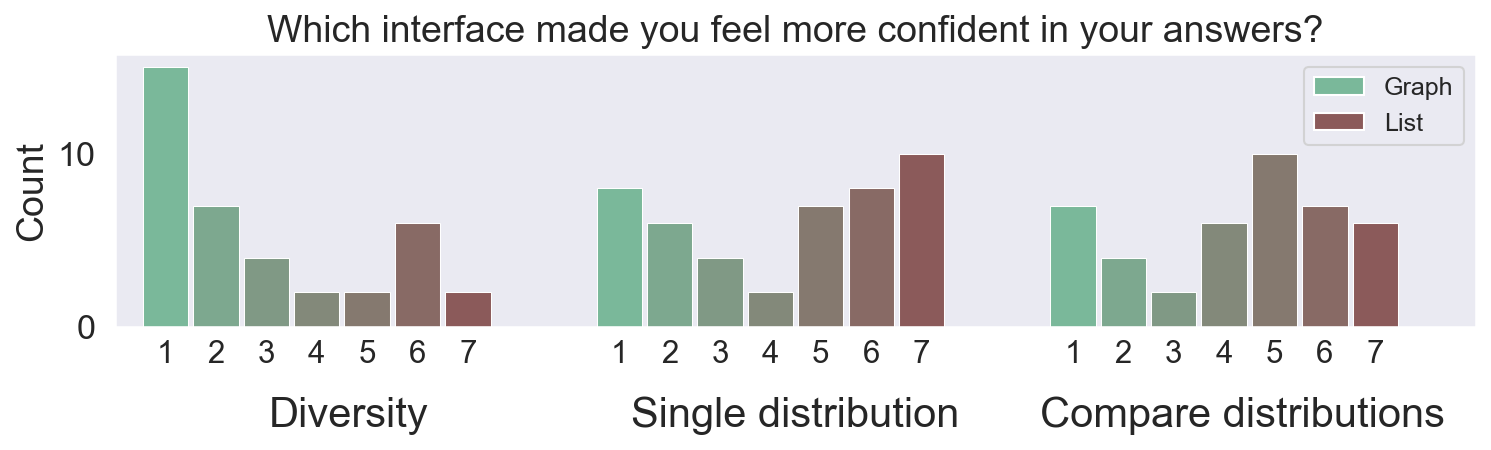}
  \end{subfigure}
  \caption{Direct comparison survey results: distribution of responses for each question (1 $=$ graph, 7 $=$ list) by study. Which interface made the task easier overall; which felt more overwhelming; which better supported understanding the range and distribution of outputs; which made participants feel more confident; overall preference. The overall preference result is also shown in the main paper (\edit{Fig.~\ref{fig:direct-comparison-preference}}).}
  \label{fig:direct-comparison-all}
\end{figure}

The survey concluded with optional free-response questions asking for thoughts (positive or negative) on each interface, including what was frustrating and what was helpful.

\begin{figure}[h]
  \centering
  \includegraphics[width=\linewidth]{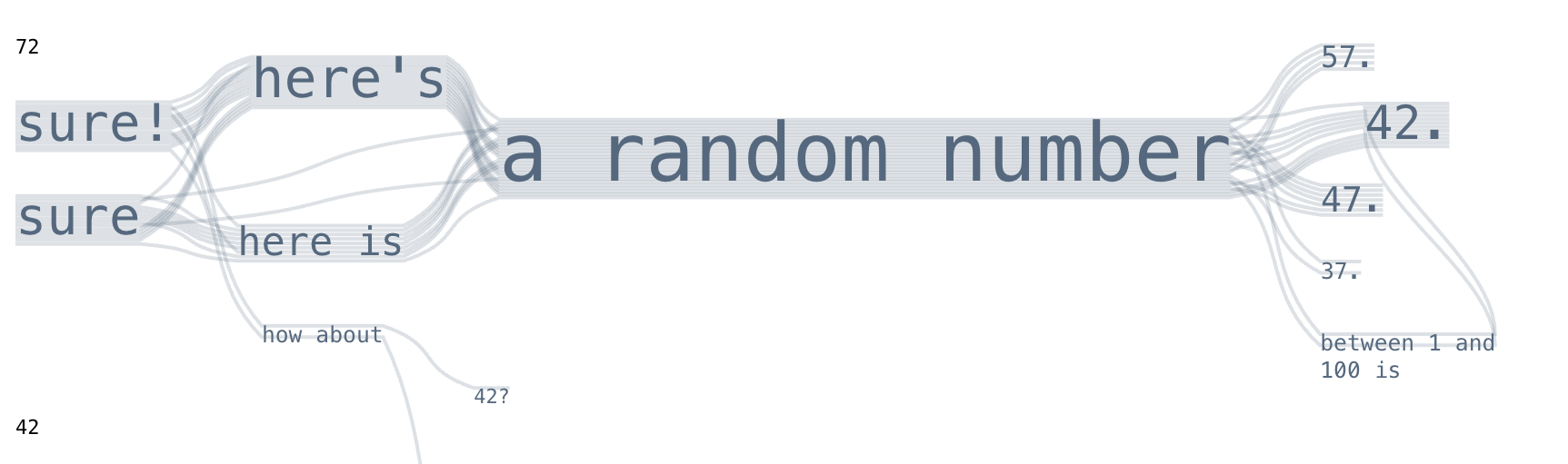}
  \caption{Outputs from the prompt ``Give me a random number between 1 and 100.'' Language models favor a small set of values (e.g., 37, 47, 57, 42); for this purely numeric task, counts or histograms are often clearer than a merged-token graph, but the graph still makes repeated modes visible (see Discussion).}
  \label{fig:random_numbers}
\end{figure}

\section{Example crowdsourced task questions}
\label{sec:example-task-questions}

We paraphrase representative accuracy items from the single-distribution and two-distribution studies. Actual survey text varied by dataset (fantasy \emph{monsters} vs.\ \emph{places}) and iteration; ground truth was always computed from the $k{=}20$ outputs each participant saw.

\subsection*{Single-distribution study}
\begin{itemize}
  \item \textbf{Most frequent entity.} ``Which creature (or location) name appears most often across the outputs?'' Scored against the argmax of manual counts in the sample.
  \item \textbf{Binned prevalence.} For several entities or traits: ``Approximately what fraction of outputs mention [X]?'' with ordered bins (0\%, 1--5\%, 6--15\%, 16--30\%, 31--50\%, and strictly more than half of outputs). Analysis used partial credit by bin distance (\S\ref{sec:eval}).
  \item \textbf{Theme.} ``Which short description best matches the majority of outputs?'' Reference themes included \emph{ethereal forest spirits or guardians guiding lost travelers} (monsters) and \emph{an ancient magical forest tied to druids or spirits} (places). We accepted exact matches after normalization or answers sharing at least two substantive words with the reference.
  \item \textbf{Recurring phrase.} ``Which option is the phrase that appears most often?'' Scored with flexible substring matching between the choice and the count-maximizing phrase.
  \item \textbf{Implausible output.} ``Which output is most likely \emph{impossible} under this prompt (even with many more samples)?'' Distractors were off-domain lines mixed with real completions.
  \item \textbf{Sample membership.} ``Which output was actually produced by this prompt among those shown?'' Correct answer was always one of the 20 displayed generations.
\end{itemize}

\subsection*{Two-distribution study}
\begin{itemize}
  \item \textbf{Relative frequency.} ``Which prompt's outputs more frequently mention [named location or creature]?'' Ground truth from paired counts in the sample.
  \item \textbf{Percentage by prompt.} ``In Prompt~1's outputs, approximately what percentage were about [named entity]?'' Binned and partial-credit scoring as in the single-distribution study.
  \item \textbf{Phrase comparison.} ``In which set do outputs more often include the phrase [\dots{}] (or similar wording)?''
  \item \textbf{Attribution.} A single sentence is shown: ``Is this output more likely from Prompt~1 or Prompt~2?'' Ground truth identified which prompt generated it.
  \item \textbf{Distributional relationship.} Items such as whether two prompts sample the ``same'' distribution, which semantic theme overlaps both sets, and which exemplar could have been produced by one prompt but not the other, with keys fixed from the displayed outputs (with prespecified relaxations when multiple labels were defensible).
\end{itemize}

%
%
%
%

\end{document}